\documentclass[runningheads]{llncs}
 \usepackage{array,multirow,graphicx}
\usepackage{amsmath,amssymb} 
\usepackage{booktabs}
\usepackage{color}
\usepackage{enumitem}
\usepackage[width=122mm,left=12mm,paperwidth=146mm,height=193mm,top=12mm,paperheight=217mm]{geometry}
\usepackage{cite} 
\usepackage[utf8]{inputenc}
\usepackage{floatrow}
\usepackage[table]{xcolor}
\usepackage{floatrow}
\usepackage{siunitx,booktabs}
\usepackage{hyperref}

\usepackage{color}

\newcommand{\scaletext}[1]{\text{\scalebox{0.7}{#1}}}

\newcommand{\myt}[2]{\mathbf{T}_\scaletext{#1}^\scaletext{#2}}
\DeclareMathOperator{\Tr}{Tr}

\newcommand\mc[1]{\multicolumn{1}{c}{#1}}

\newcolumntype{d}[1]{D{.}{.}{#1}}
\newcolumntype{R}{@{\extracolsep{3cm}}r@{\extracolsep{0pt}}}
\newcommand{\boldentry}[2]{
  \multicolumn{1}{S[table-format=#1,
                    mode=text,
                    text-rm=\bfseries\selectfont
                   ]}{#2}}
\newcommand{\sizeA}{5.4}
\newcommand{\sizeB}{2.2}
\newcommand{\sizeC}{2.2}

\begin{document}

\pagestyle{headings}
\mainmatter
\def\ECCV18SubNumber{982}  

\title{A Framework for Evaluating \\ 6-DOF Object Trackers} 

\titlerunning{A Framework for Evaluating 6-DOF Object Trackers}

\authorrunning{Mathieu Garon, Denis Laurendeau and Jean-François Lalonde}

\author{Mathieu Garon \orcidID{0000-0003-1811-4156}, Denis Laurendeau \orcidID{0000-0003-2858-5955} and Jean-François Lalonde \orcidID{0000-0002-6583-2364}}
\institute{Université Laval\footnote{\scriptsize \texttt{mathieu.garon.2@ulaval.ca, denis.laurendeau@gel.ulaval.ca, jflalonde@gel.ulaval.ca}}}

\maketitle

\begin{abstract}
We present a challenging and realistic novel dataset for evaluating 6-DOF object tracking algorithms. Existing datasets show serious limitations---notably, unrealistic synthetic data, or real data with large fiducial markers---preventing the community from obtaining an accurate picture of the state-of-the-art. Using a data acquisition pipeline based on a commercial motion capture system for acquiring accurate ground truth poses of real objects with respect to a Kinect V2 camera, we build a dataset which contains a total of 297 calibrated sequences. They are acquired in three different scenarios to evaluate the performance of trackers: \emph{stability}, robustness to \emph{occlusion} and accuracy during challenging \emph{interactions} between a person and the object. We conduct an extensive study of a deep 6-DOF tracking architecture and determine a set of optimal parameters. We enhance the architecture and the training methodology to train a 6-DOF tracker that can robustly generalize to objects never seen during training, and demonstrate favorable performance compared to previous approaches trained specifically on the objects to track. 
\end{abstract}
\section{Introduction}
\label{sec:introduction}



With the recent emergence of 3D-enabled augmented reality devices, tracking 3D objects in 6 degrees of freedom (DOF) is a problem that has received increased attention in the past few years. As opposed to SLAM-based camera localization techniques---now robustly implemented on-board various commercial devices---that can use features from the entire scene, 6-DOF object tracking approaches have to rely on features present on a (typically small) object, making it a challenging problem. Despite this, recent approaches have demonstrated tremendous performance both in terms of speed and accuracy~\cite{kehl2017real,tan2017looking,garon-tvcg-17}. 

\begin{figure}[t]
\centering
\scriptsize
\setlength{\tabcolsep}{8pt}
\begin{tabular}{ccc}
\includegraphics[height=2.2cm]{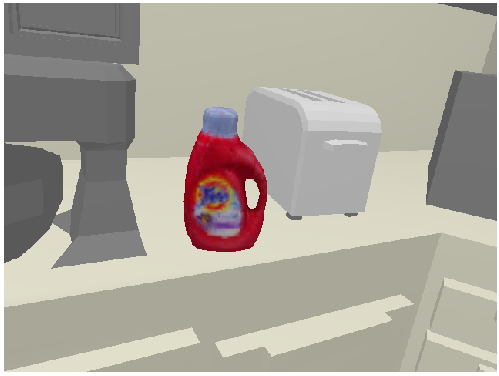} & 
\includegraphics[trim={7cm 3cm 7cm 3cm},clip,height=2.2cm]{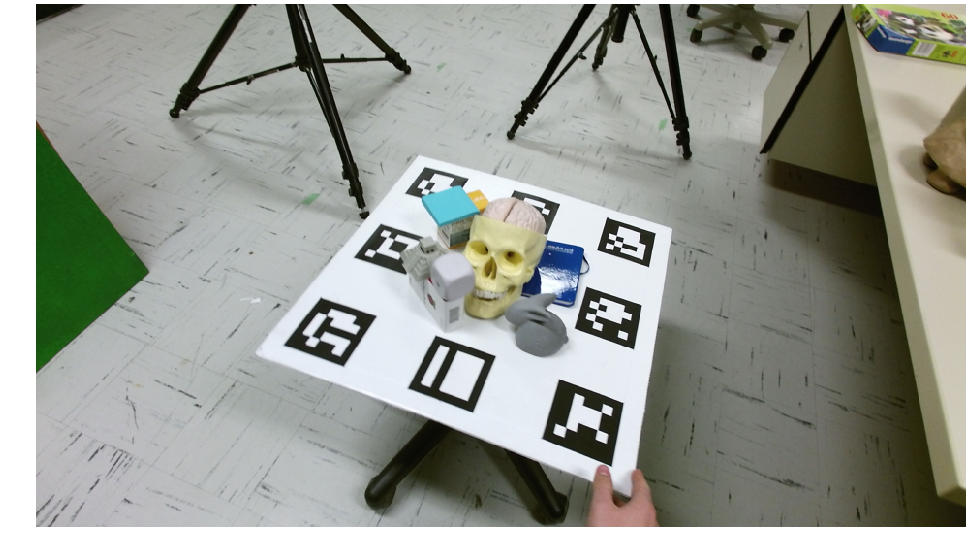} &
\includegraphics[trim={11cm 10cm 13cm 1.2cm},clip,height=2.2cm]{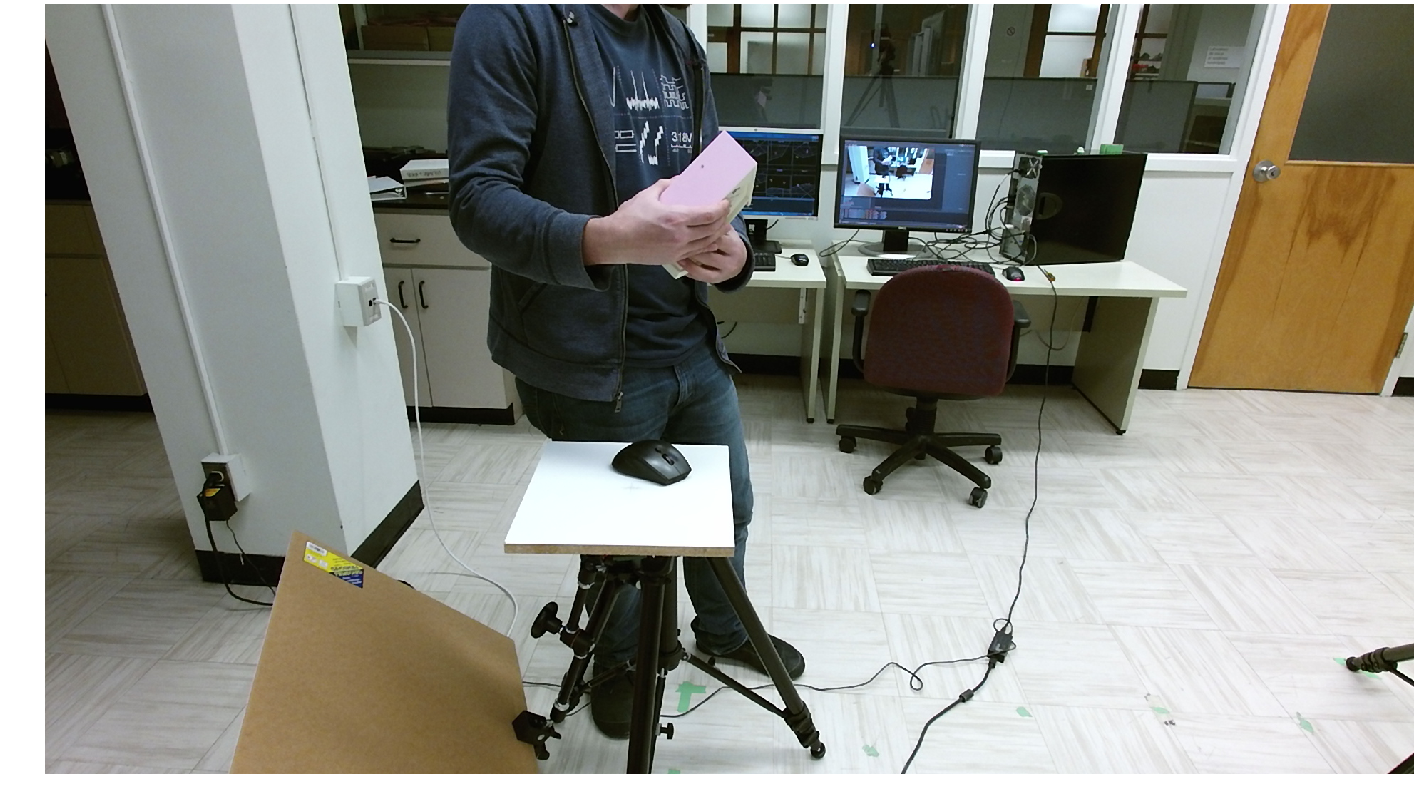} \\
\includegraphics[height=2.2cm]{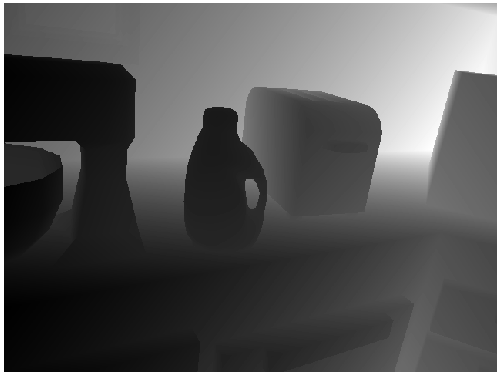} & 
\includegraphics[trim={7cm 3cm 7cm 3cm},clip,height=2.2cm]{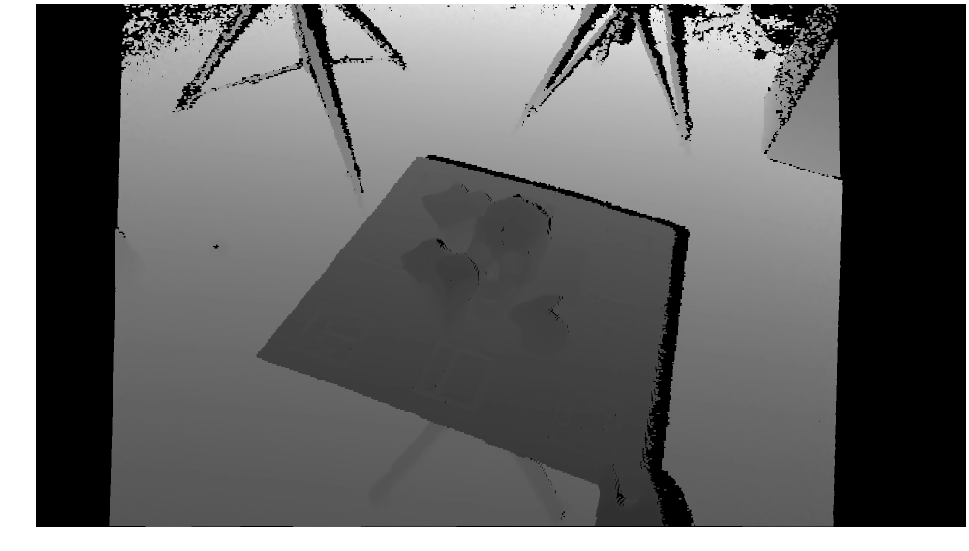} &
\includegraphics[trim={11cm 10cm 13cm 1.2cm},clip,height=2.2cm]{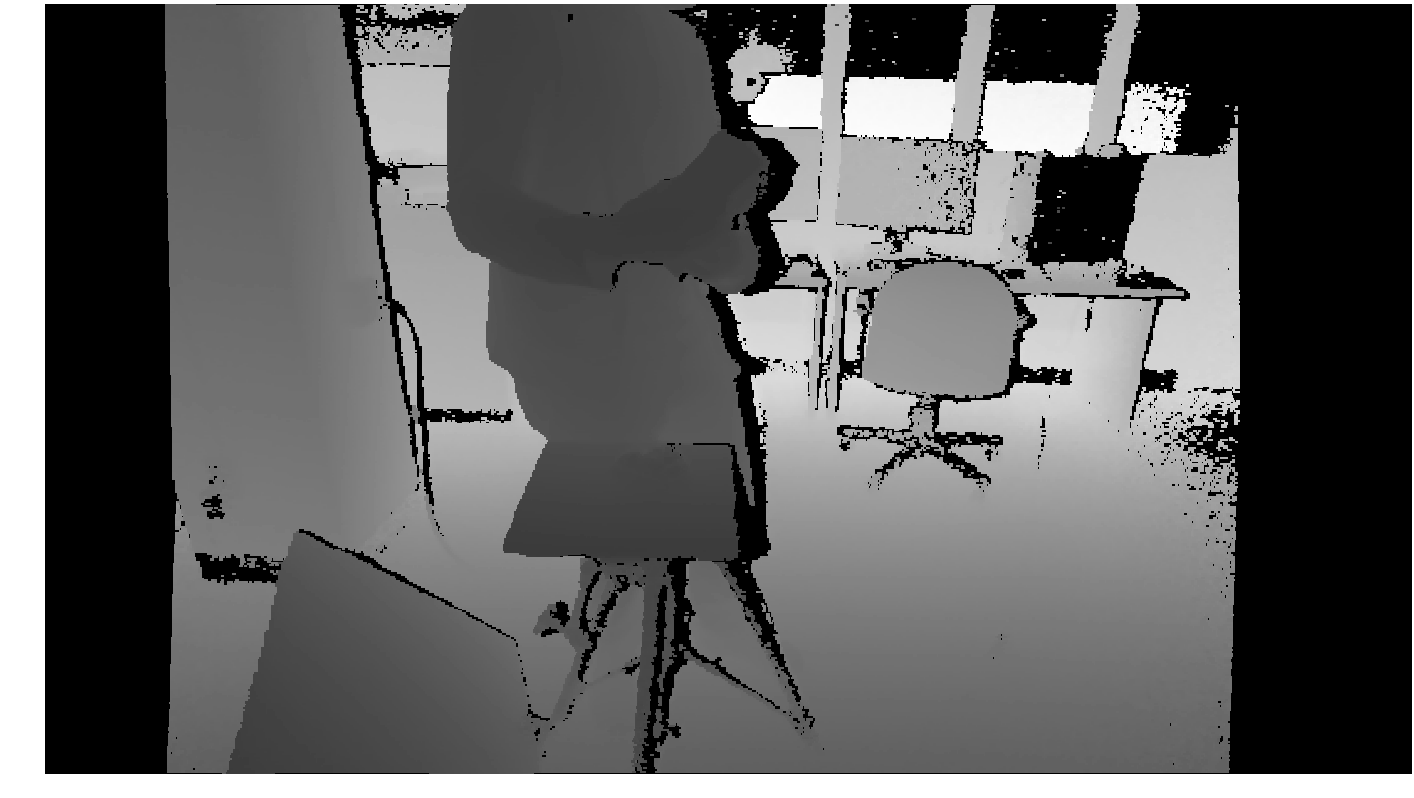} \\
(a) Choi-Christensen~\cite{choi2013rgb} & (b) Garon-Lalonde~\cite{garon-tvcg-17} & (c) Ours
\end{tabular}
\caption[]{Comparison of datasets for evaluating 6-DOF tracking algorithms. Typical RGB (top) and depth (bottom) frames for (a) the synthetic dataset of Choi and Christensen~\cite{choi2013rgb}, (b) the real dataset of Garon and Lalonde~\cite{garon-tvcg-17}, and (c) ours. Compared to the previous work, our dataset contains real objects captured by a sensor, and does not use a calibration board, therefore mimicking realistic real-world scenarios. } 
\label{fig:introduction}
\end{figure}

Unfortunately, obtaining an accurate assessment of the performance of 6-DOF object tracking approaches is becoming increasingly difficult since accuracy on the main dataset used for this purpose has now reached closed to 100\%. Introduced in 2013 by Choi and Christensen~\cite{choi2013rgb}, their dataset consists of 4 short sequences of purely synthetic scenes. The scenes are made of unrealistic, texture-less backgrounds with a single colored object to track, resulting in noiseless RGBD images (see fig.~\ref{fig:introduction}-(a)). The object is static and the camera rotates around it in wide motions, occasionally creating small occlusions (at most 20\% of the object is occluded). While challenging at first, the dataset has now essentially been solved for the RGBD case. For example, the method of Kehl et al.~\cite{kehl2017real} (2017) reports an average error in translation/rotation of 0.5mm/$0.26^\circ$, which is an improvement of 0.3mm/$0.1^\circ$ over the work of Tan et al. (2015)~\cite{tan-iccv-15}, who have themselves reported a 0.01mm/$1^\circ$ improvement to the approach designed by Krull et al. (2014)~\cite{krull20146}. The state of the art on the dataset has reached a near-perfect error of 0.1mm/$0.07^\circ$~\cite{tan2017looking}, which highlights the need for a new dataset with more challenging scenarios.


Another dataset, introduced by Garon and Lalonde~\cite{garon-tvcg-17}, includes 12 sequences of real objects captured with real sensors. While a significant improvement over the synthetic dataset of~\cite{choi2013rgb}, dealing with real data raises the issue of providing accurate ground truth pose of the object at all times. To obtain this ground truth information, their strategy (also adopted in 6-DOF \emph{detection} datasets~\cite{hinterstoisser2012model,hodan2017t}) is to use calibration boards with fiducial markers. While useful to accurately and easily determine an object pose, this has the unfortunate consequence of constraining the object to lie on a large planar surface (fig.~\ref{fig:introduction}-(b)). 

In this paper, we present a novel dataset allowing the systematic evaluation of 6-DOF tracking algorithms in a wide variety of real scenarios without requiring calibration boards (fig.~\ref{fig:introduction}-(c)). Our dataset is one order of magnitude larger than the previous work: it contains 297 sequences of 11 real objects. The sequences are split into 3 different scenarios, which we refer to as \emph{stability}, \emph{occlusion}, and \emph{interaction}. The \emph{stability} scenario aims at quantifying the degree of jitter in a tracker. The object is kept static and placed at various angles and distances from the camera. The \emph{occlusion} scenario, inspired by \cite{garon-tvcg-17}, has the object rotating on a turntable and being progressively occluded by a flat panel. Occlusion ranges from 0\% (unoccluded) to 75\%, thereby testing trackers in very challenging situations. Finally, in the \emph{interaction} scenario, a person is moving the object around freely in front of the camera (fig.~\ref{fig:introduction}-(c)), creating occlusions and varying object speed. 

In addition, we also introduce two new 6-DOF real-time object trackers based on deep learning. The first, trained for a specific object, achieves state-of-the-art performance on the new dataset. The second, trained \emph{without a priori knowledge of the object to track}, is able to achieve an accuracy that is comparable with previous work trained specifically on the object. These two trackers rely on the same deep learning architecture and only differ in the training data. Furthermore, both of our trackers have the additional significant advantage of requiring only \emph{synthetic} training data (i.e. no real data is needed for training). We believe this is an exciting first step in the direction of training generic trackers which do not require knowledge of the object to track at training time. 

In summary, this paper brings 3 key contributions to 6-DOF object tracking: 
\begin{enumerate}[noitemsep]
\item A novel dataset of real RGBD sequences for the systematic evaluation of 6-DOF tracking algorithms that is one order of magnitude larger than existing ones, and contains 3 challenging scenarios; 
\item A real-time deep learning architecture for tracking objects in 6-DOF which is more stable and more robust to occlusions than previous approaches; 
\item A generic 6-DOF object tracker trained without knowledge of the object to track, achieving performance on par with previous approaches trained specifically on the object. 
\end{enumerate}


\section{Related work}
\label{sec:related_work}

There are two main relevant aspects in 6-DOF pose estimation: single frame object detection and multi-frame temporal trackers. The former has received a lot of attention in the literature and benefits from a large range of public datasets. The most notorious dataset is arguably \emph{Linemod}~\cite{hinterstoisser2012model}, which provide 15 objects with their mesh models and surface colors. To obtain the ground truth object pose, a calibration board with fiducial markers is used. Since then, many authors created similar but more challenging benchmarks~\cite{tejani2014latent,doumanoglou2016recovering,hodan2017t}. However these datasets do not contain temporal and displacement correlation between each frame, which makes them inadequate for evaluating temporal trackers. 


In the case of temporal tracking, only a few datasets exist to evaluate the approaches. As mentioned in the introduction, the current, widely used standard dataset is the synthetic dataset of Choi and Christensen~\cite{choi2013rgb}, which contains 4 sequences with 4 objects rendered in a texture-less virtual scene. Another available option is the one provided by Akkaladevi et al.~\cite{akkaladevi2016tracking} who captured a single sequence of a scene containing 4 different objects with a Primesense sensor. However, the 3D models are not complete and do not include training data that could be exploited by learning-based methods. Finally, recent work by Garon and Lalonde~\cite{garon-tvcg-17} proposed a public dataset of 4 objects containing 4 sequences with clutter and an additional set of 8 sequences with controlled occlusion on a specific object. Fiducial markers are used to generate the ground truth pose of the model, which limits the range of displacements that can be achieved. In contrast, we propose a new method to collect ground truth pose data that makes the acquisition simpler without the need for fiducial markers. 

There is an increasing interest in 6-DOF temporal trackers since they were shown to be faster and more robust than single frame detection methods. In the past, geometric methods based on ICP~\cite{choi2013rgb,aldoma2013multimodal,kwon2007particle,chitchian2013adapting} were used for temporal tracking, but they lack robustness for small objects and are generally computationally expensive. Data-driven approaches such as the ones reported in \cite{krull20146,tan-iccv-15,tan2014multi} can learn more robust features and the use of the Random Forest regressor~\cite{breiman2001random} decreases the computing overhead significantly. Other methods show that the contours of the objects in RGB and depth data provide important cues for estimating pose~\cite{tan2017looking,kehl2017real,tjaden2017real}. While their optimization techniques can be accurate, many assumptions are made on the features which restrict the type of object and the type of background that can be dealt with. Recently, Garon and Lalonde~\cite{garon-tvcg-17} proposed a deep learning framework which can learn robust features automatically from data. They use a feedback loop by rendering the 3D model at runtime at the previous pose, and regress the pose difference between the rendered object and the real image. While their method compares to the previous work with respect to accuracy, their learned features are more robust to higher level of occlusion and noise. A downside is that their method needs a dataset of real images and a specific network has to be trained for each object which can be time consuming. We take advantage of their architecture but introduce novel ideas to provide a better performing tracker that can be trained entirely on synthetic data. In addition, our network can be trained to generalize to previously unseen objects.

\section{Dataset capture and calibration}
\label{sec:dataset_capture}

Building a dataset with calibrated object pose w.r.t the sensor at each frame is a challenging task since it requires an accurate method to collect ground truth object pose. Until now, the most practical method to achieve this task was to use fiducial markers and calibrate the object pose w.r.t these markers~\cite{garon-tvcg-17,hinterstoisser2012model,hodan2017t,tejani2014latent}. However, this method suffers from two main drawbacks. First, the object cannot be moved independently of the panel so this restricts the camera to move around the object of interest. Second, the scene always contains visual cues (the markers) which could involuntarily ``help'' the algorithms. 

Our approach eliminates these limitations. A Vicon\texttrademark~ MX-T40 motion capture system is used to collect the ground truth pose of the objects in the scene. The retroreflective Vicon markers that must be used are very small in size (3mm diameter) and can be automatically removed in a post-processing step. In this section, we describe the capture setup and the various calibration steps needed to align the object model and estimate its ground truth pose. The resulting RGBD video sequences captured using this setup are presented in sec.~\ref{sec:evaluation_methodology}. 


\subsection{Capture setup}

\begin{figure}[!t]
\centering
\footnotesize
\setlength{\tabcolsep}{.2cm}
\begin{tabular}{cc}
\includegraphics[height=4cm]{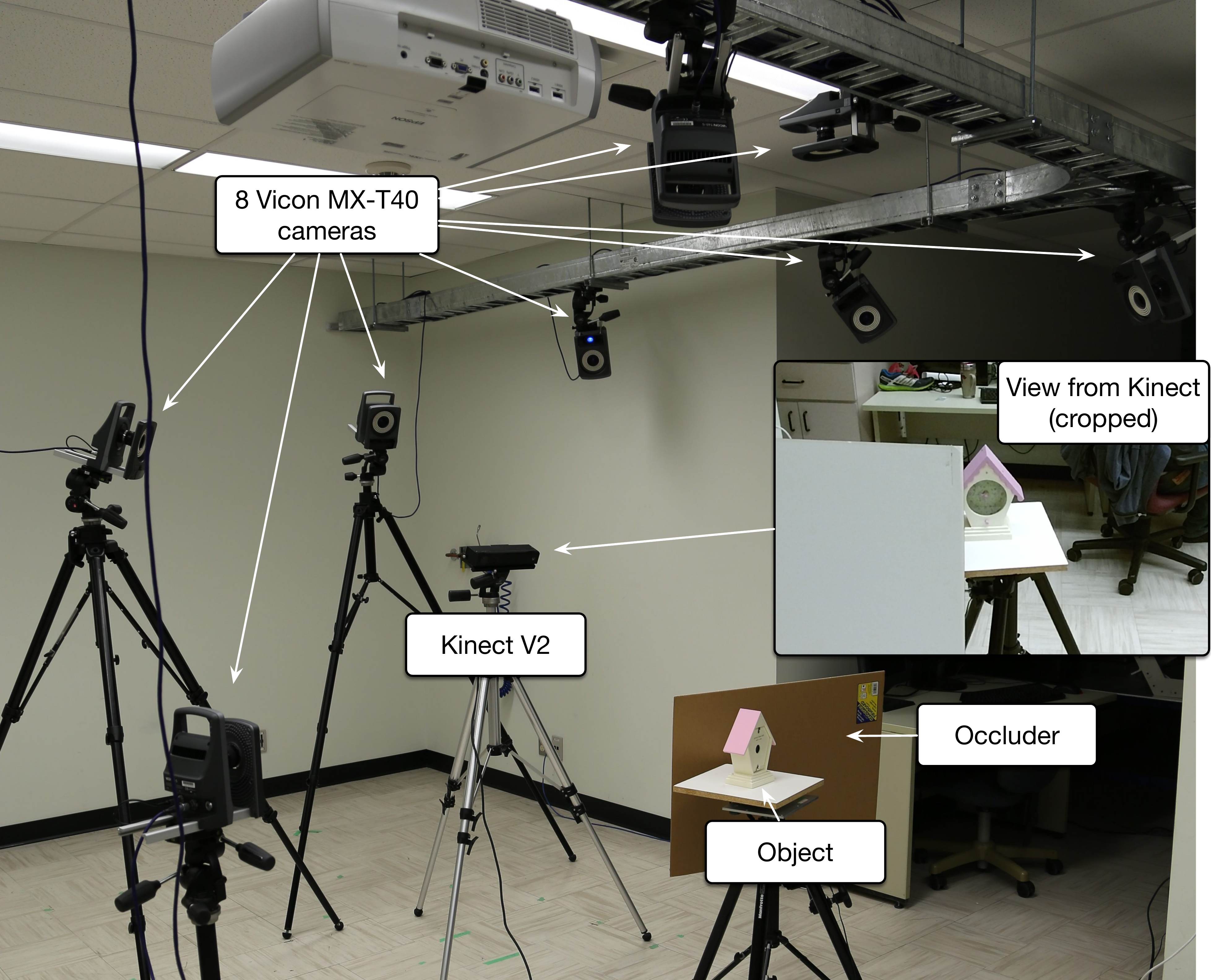} & 
\includegraphics[height=4cm]{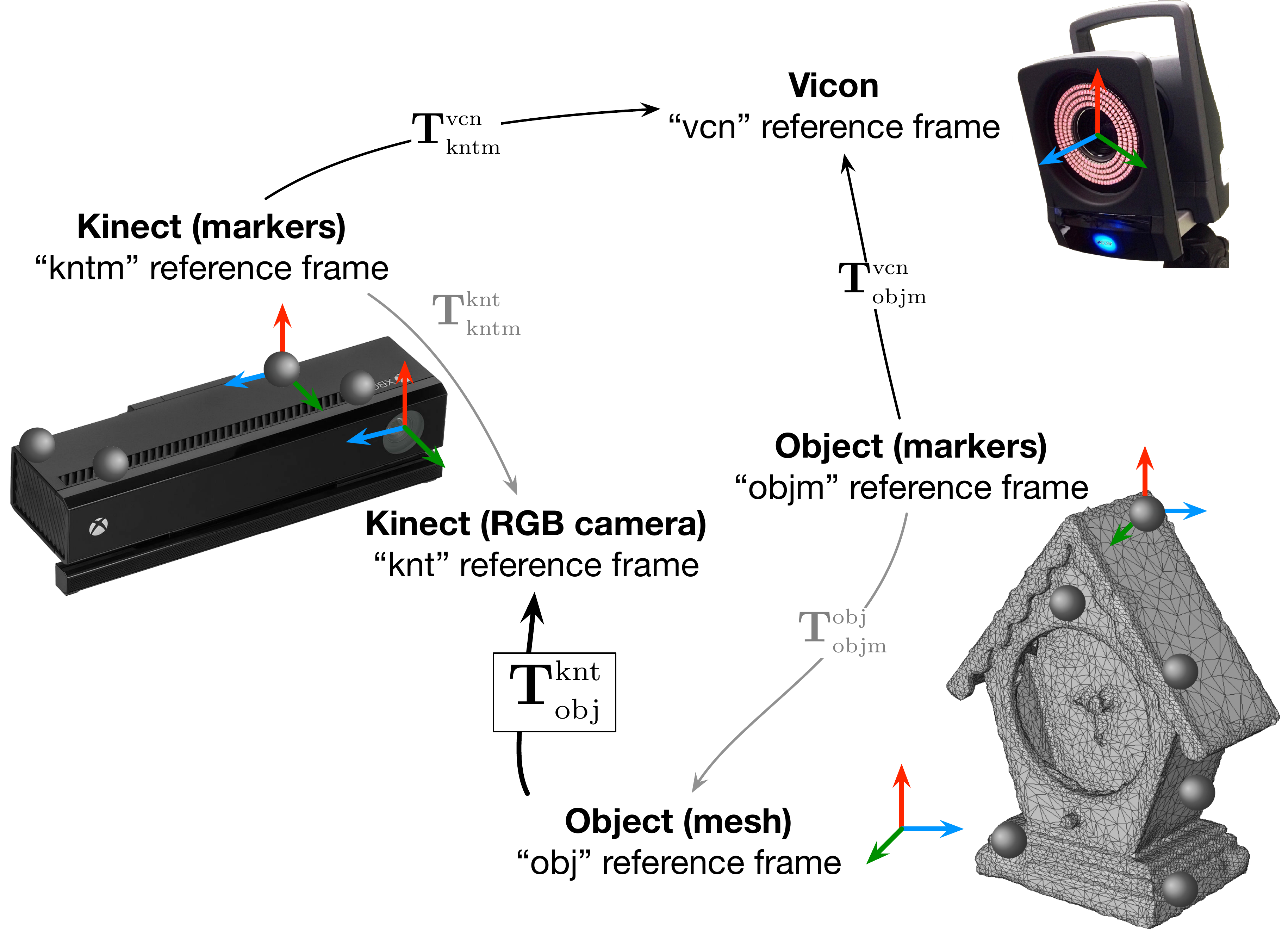} \\
(a) & (b) 
\end{tabular}
\caption[]{Acquisition setup used to capture our novel dataset. (a) Actual setup, which includes an 8-camera Vicon motion capture system and a Kinect V2. The resulting view from the Kinect is shown in the inset. Here, an occluder is placed in front of the object. (b) The various transformations that must be calibrated in order to obtain the object pose in the Kinect RGB camera reference frame $\myt{obj}{knt}$. The transformations shown in black are obtained from the motion capture system directly, while the gray ones need a specific calibration procedure described in the main body of the paper.}
\label{fig:capture-setup}
\end{figure}

The motion capture setup is composed of a set of 8 calibrated cameras that track retro-reflective markers of 3mm in diameter installed on the objects of interest in a $3\times3\times3\text{m}^3$ work area. Vicon systems can provide a marker detection accuracy of up to 0.15 mm on static objects and 2mm on moving objects according to \cite{merriaux2017study}. A Kinect V2 is used to acquire the RGBD frames, and is calibrated with the Vicon to record the ground truth pose of the objects in the Kinect coordinate system. The actual setup used to capture the dataset is shown in fig.~\ref{fig:capture-setup}-(a). 

\subsection{Calibration}
\label{subsec:calibration}

With an RGB-D sensor such as the Kinect V2, color and depth values are projected onto two different planes. We define the Kinect reference frame (``knt'') as the origin of its RGB camera, and align the depth data by reprojecting it to the color plane using the factory calibration parameters. We calibrate the depth correction as in Hodan et al.~\cite{hodan2017t}. In this section, the notation $\myt{a}{b}$ is used to denote a rigid transformation from reference frame ``a'' to ``b''.

We aim to recover the pose of the object in the Kinect reference frame $\myt{obj}{knt}$ (fig.~\ref{fig:capture-setup}-(b)). To do so, we first rely on the Vicon motion capture system, which has its own reference frame ``vcn''. The set of retroreflective markers installed on the object define the local reference frame ``objm''. Similarly, the set of markers placed on the Kinect define the local reference frame ``kntm''. The Vicon system provides the transformations $\myt{objm}{vcn}$ and $\myt{kntm}{vcn}$ directly, that is, the mappings between the object and Kinect markers and the Vicon reference frame respectively. The transformation between the object markers and the Kinect markers is obtained by chaining the previous transformations:
\begin{equation} 
\label{eq:marker-transformation}
\myt{objm}{kntm} = (\myt{kntm}{vcn})^{-1} \myt{objm}{vcn} \,.
\end{equation}
The pose $\myt{obj}{knt}$ is recovered with the transformations between the local frames defined by the markers and the object/Kinect reference frames $\myt{obj}{objm}$ and $\myt{knt}{kntm}$: 
\begin{equation} 
\label{eq:obj-transformation}
\myt{obj}{knt} = (\myt{knt}{kntm})^{-1} \myt{objm}{kntm} \myt{obj}{objm} \,.
\end{equation}
The calibration procedures needed to obtain these two transformations, also shown in gray in fig.~\ref{fig:capture-setup}-(b), are detailed next. 

\subsubsection{Kinect calibration}

In order to find the transformation $\myt{kntm}{knt}$ between the local frame defined by the markers installed on the Kinect and its RGB camera, we rely on a planar checkerboard target on which Vicon markers are randomly placed. Then, the position of each corner of the checkerboard is determined with respect to the markers with the following procedure. A 15cm-long pen-like probe that has a 1cm Vicon marker attached at one end was designed for this purpose. The sharp end is placed on the corner to be detected, and the probe is moved in a circular motion around that point. A sphere is then fit (using least-squares) to the resulting marker positions (achieving an average radius estimation error of 0.7mm), and the center of the sphere is kept as the location for the checkerboard corner. The checkerboard target was then moved in the capture volume and corners were detected by the Kinect RGB camera, thereby establishing 2D-3D correspondences between these points. The perspective-$n$-points algorithm~\cite{zhang2000flexible} was finally used to compute $\myt{kntm}{knt}$. 

\subsubsection{Object calibration}

To estimate the transformation between the local frame defined by the markers placed on the object and its mesh coordinate system $\myt{objm}{obj}$, we rely on the Kinect pose calibrated with the method described previously. As a convention, we define the origin of the object local coordinate system at the center of mass of the markers, the same convention is used for the mesh by using the center of mass of the vertices. We roughly align the axis and use ICP to refine its position (based on the Kinect depth values). Finally, with the help of a visual interface where a user can move and visualize the aligment of the object, fine-scale adjustments can be performed manually from several viewpoints to minimize the error between the observed object and the reprojected mesh.


\subsubsection{Synchronization}
\label{subsec:synchronisation}

In addition to spatial calibration, precise temporal alignment must be achieved to synchronize Vicon and Kinect frames. Unfortunately, the Kinect does not offer hardware synchronization capabilities, therefore we adopt the following software solution. We assume that the sequences are short enough to neglect clock drift. We also assume a stable sampling of the Vicon system on a high bandwidth closed network. In this setup, synchronization can be achieved by estimating the (constant) time difference $\delta t$ between the Vicon and the Kinect frame timestamps. By moving the checkboard of sec.\ref{subsec:calibration} with varying speed, we estimate the $\delta t$ that minimizes the reprojection error between the checkerboard corners from sec.~\ref{subsec:calibration} and the Vicon markers.

\subsubsection{Removing the markers}
\label{subsec:markers}

\begin{figure}[!t]
\centering
\footnotesize
\setlength{\tabcolsep}{.2cm}
\includegraphics[width=\linewidth]{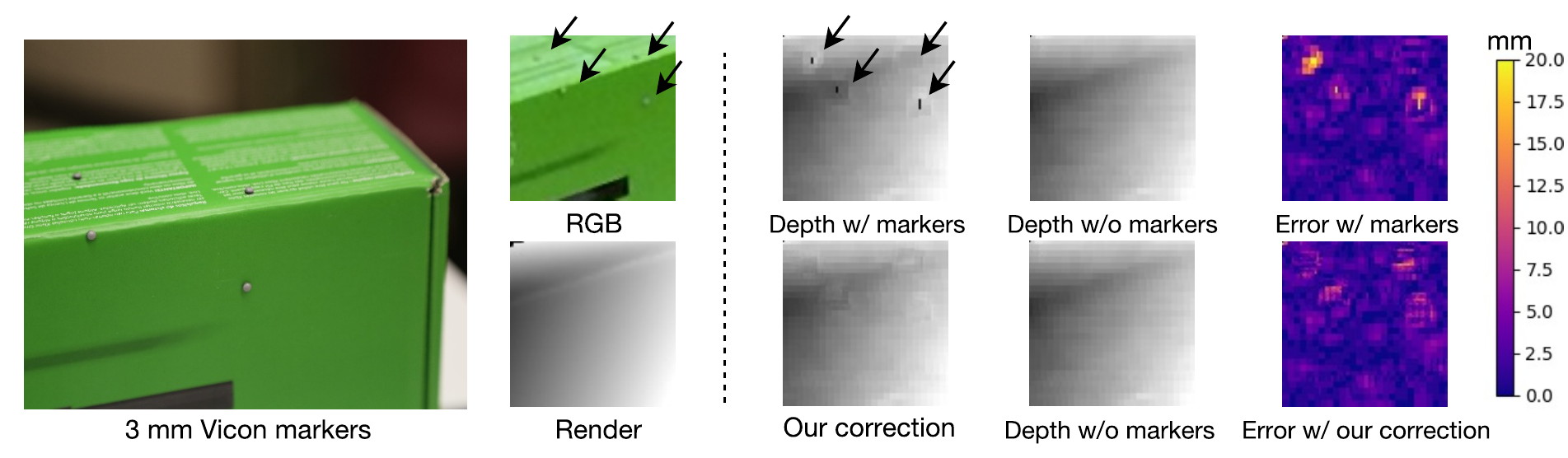}
\caption[]{ Example of an RGB and depth frame containing 2 markers on a flat surface, and 2 markers near an edge. We take advantage of our knowledge of the object mesh and pose to replace patches of $10\times10$ pixels around the marker by the depth values of a render at the same pose. We capture an image without the markers to compare the error. On the modified patches we report a RMSE of 139.8 mm on the depth with the markers, and 4.7 with the corrected version.}
\label{fig:markers}
\end{figure}

The 3mm markers used to track the object are retro-reflective and, despite their small size and their low number (7 per object on average), they nevertheless create visible artifacts in the depth data measured by the Kinect, see fig.~\ref{fig:markers}. We propose a post-processing algorithm for automatically removing them in all the sequences. First, to ensure that the marker can be observed by the Kinect we reproject the (known) marker positions onto the depth image and compute the median distance between the depth in a small window around the reprojected point and its ground truth depth. If the difference is less than 1cm, the point is considered as not occluded, and will be processed. Finally, we render the depth values of the 3D model at the given pose and replace the $10\times10$ pixel window from the original image with the rendered depth values. For more realism, a small amount of gaussian noise is added. Pixels from the background are simply ignored. On average, only 3.4\% of the object pixels are corrected. We also minimize the chances of affecting the geometric structure of the object by placing the markers on planar surfaces. Fig.~\ref{fig:markers} shows a comparison of the error between a Kinect depth image captured with markers, and another image of the same scene with markers that have been corrected with our algorithm. The RMSE of the pixel patches around the markers is 139.8 mm without the correction, and 4.7 mm with the correction.






\section{Dataset scenarios, metrics, and statistics}
\label{sec:evaluation_methodology}

This section defines novel ways to evaluate 6-DOF trackers using calibrated sequences captured with the setup presented in sec.~\ref{sec:dataset_capture}. We provide an evaluation methodology that will reflect the overall performance of a tracker in different scenarios. To attain this objective, we captured 297 sequences of 11 different objects of various shapes in 3 scenarios: \emph{stability}, \emph{occlusion}, and \emph{interaction}. We also provide quantitative metrics to measure the performance in each scenario. Our dataset and accompanying code is available at \url{http://www.jflalonde.ca/projects/6dofObjectTracking}. 


\subsection{Performance metrics}

Before we describe each scenario, we first introduce how we propose to evaluate the difference between two poses $\mathbf{P}_1$ and $\mathbf{P}_2$. Here, a pose $\mathbf{P} = \left[\begin{array}{cc} \mathbf{R} & \mathbf{t} \end{array} \right]$ is described by a rotation matrix $\mathbf{R}$ and a translation vector $\mathbf{t}$. Previous works consider the average of each axis component in translation and rotation separately. The side effect of this metric is that a large error on a single component is less penalized. To overcome this limitation, the translation error is simply defined as the L2 norm between the two translation vectors: 
\begin{equation}
\delta_\mathbf{t}(\mathbf{t}_1, \mathbf{t}_2) = ||\mathbf{t_1} - \mathbf{t_2}||_2 \,. 
\end{equation}
The distance between two rotation matrices is computed using:
\begin{equation}
\delta_\mathbf{R}(\mathbf{R}_1, \mathbf{R}_2) = \arccos \left(\frac{\Tr (\mathbf{R}_{1}^T\mathbf{R}_2) - 1}{2} \right) \,,
\end{equation}
where $\Tr(\cdot)$ denotes the matrix trace.

\subsection{Scenarios}
\subsubsection{The \emph{stability} scenario}
\label{subsec:stability}

In this first scenario, we propose to quantify the degree of pose jitter when tracking a static object. To evaluate this, we captured 5-second sequences of the object under 4 different viewpoints and with 3 configurations: at a distance of 0.8m from the sensor (``near''), of 1.5m from the sensor (``far''), and of 0.8m from the sensor, but this time with distractor objects partly occluding the object of interest (``occluded'').
To measure the stability, Tan et al.~\cite{tan2017looking} use the standard deviation of the pose parameters on a sequence. We propose a different metric inspired from \cite{niehorster2017accuracy} that penalizes variation from frame to frame instead of the general distribution across the sequence. We compute the distance between poses $\mathbf{P}_{i-1}$ and $\mathbf{P}_{i}$ at time $i$. In other words, we report the distribution of $\delta_\mathbf{t}(\mathbf{t}_{i-1}, \mathbf{t}_i)$ and ${\delta_\mathbf{R}(\mathbf{R}_{i-1}, \mathbf{R}_i)}$ for all frames of the stability scenario.

\subsubsection{The \emph{occlusion} scenario}
\label{subsec:occlusion}

To evaluate the robustness to occlusion, we follow \cite{garon-tvcg-17} and place the object on a turntable at 1.2m from the sensor, and a static occluder is placed in front of the object in a vertical and horizontal position. We compute the amount of occlusion based on the largest dimension of the object, and provide sequences for each object from 0\% to 75\% occlusion in 15\% increments, which results in a total of 11 sequences per object. 
Here, we compute errors by comparing the pose $\mathbf{P}_i$ at time $i$ with the ground truth $\mathbf{P}^*_i$ for that same frame, i.e., $\delta_\mathbf{t}(\mathbf{t}^*_i, \mathbf{t}_i)$ and $\delta_\mathbf{R}(\mathbf{R}^*_i, \mathbf{R}_i)$. Temporal trackers may lose tracking on difficult frames. This can affect the overall score depending on the moment where the tracker fails. To bypass this limitation, we initialize the tracker at the ground truth pose $\mathbf{P}^*_i$ every 15 frames as in \cite{garon-tvcg-17}. 

\subsubsection{The \emph{interaction} scenario}
\label{subsec:free-form_interaction}


In this last scenario, the experimenter holds the object in his hands and manipulates it in 4 different ways: 1) by moving the object around but without rotating it (``translation-only''); 2) by rotating the object on itself without translating it (``rotation-only''); 3) by freely moving and rotating the object around at low speeds (``free-slow''); and 4) by freely moving and rotating the object at higher speeds and by voluntarily generating more occlusions (``free-hard''). In all situations but the ``free-hard'', we reset the tracker every 15 frames and we report $\delta_\mathbf{t}(\mathbf{t}^*_i, \mathbf{t}_i)$ and $\delta_\mathbf{R}(\mathbf{R}^*_i, \mathbf{R}_i)$ as in sec.~\ref{subsec:occlusion}. 
Since the object speed varies, we also compute the translational and rotational inter-frame displacement ($\delta_\mathbf{t}(\mathbf{t}^*_{i-1}, \mathbf{t}^*_i)$, $\delta_\mathbf{R}(\mathbf{R}^*_{i-1}, \mathbf{R}^*_i)$) and report the performance metric above as a function of that displacement. In addition, it is also informative to count the number of times the tracker has failed. We consider a tracking failure when either $\delta_\mathbf{t}(\mathbf{t}^*_i, \mathbf{t}_i)> 3 \text{cm}$ or $\delta_\mathbf{R}(\mathbf{R}^*_i, \mathbf{R}_i) > 20^\circ$ for more than 7 consecutive frames. When a failure is detected, the tracker is reset at the ground truth pose $\mathbf{P}^*$. We report these failures on the ``free-hard'' sequences only. 

\begin{figure}[t!]
\centering
\tiny
\setlength{\tabcolsep}{1pt}
\begin{tabular}{cccccccc}
clock (222) & dragon (207) & dog (187) & shoe (314) & kinect (287) & skull (218) & turtle (225) \\
\includegraphics[trim=2cm 1.5cm 2cm 1.5cm,clip,width=.13\linewidth]{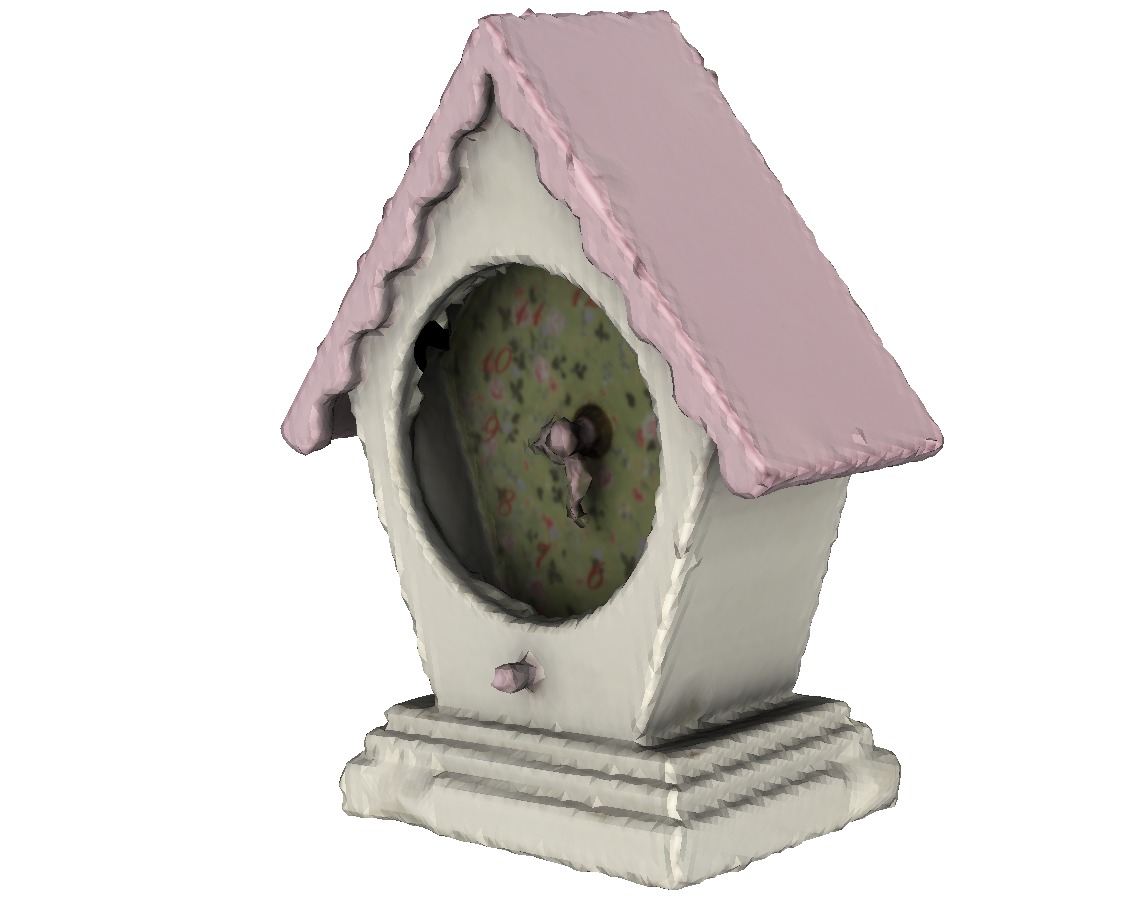} &
\includegraphics[trim=2cm 1.5cm 2cm 1.5cm,clip,width=.13\linewidth]{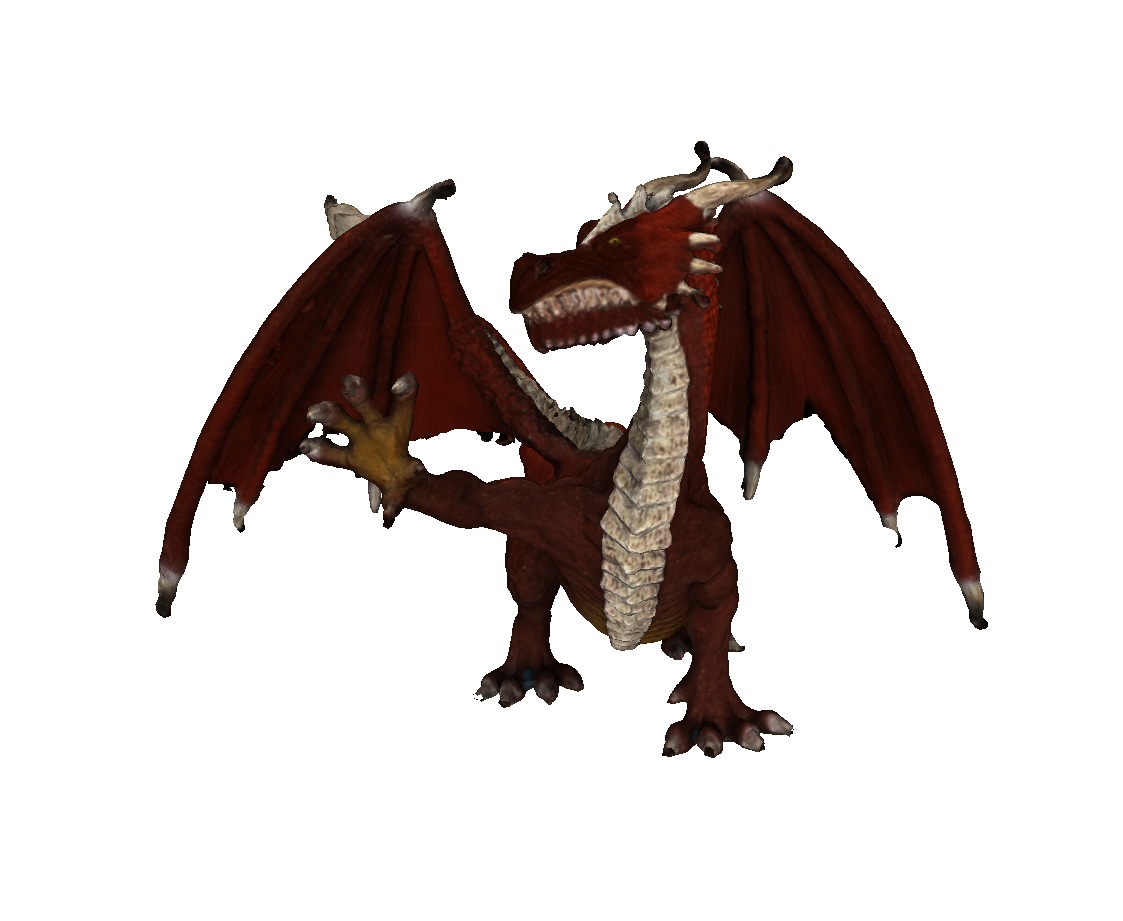} &
\includegraphics[trim=2cm 1.5cm 2cm 1.5cm,clip,width=.13\linewidth]{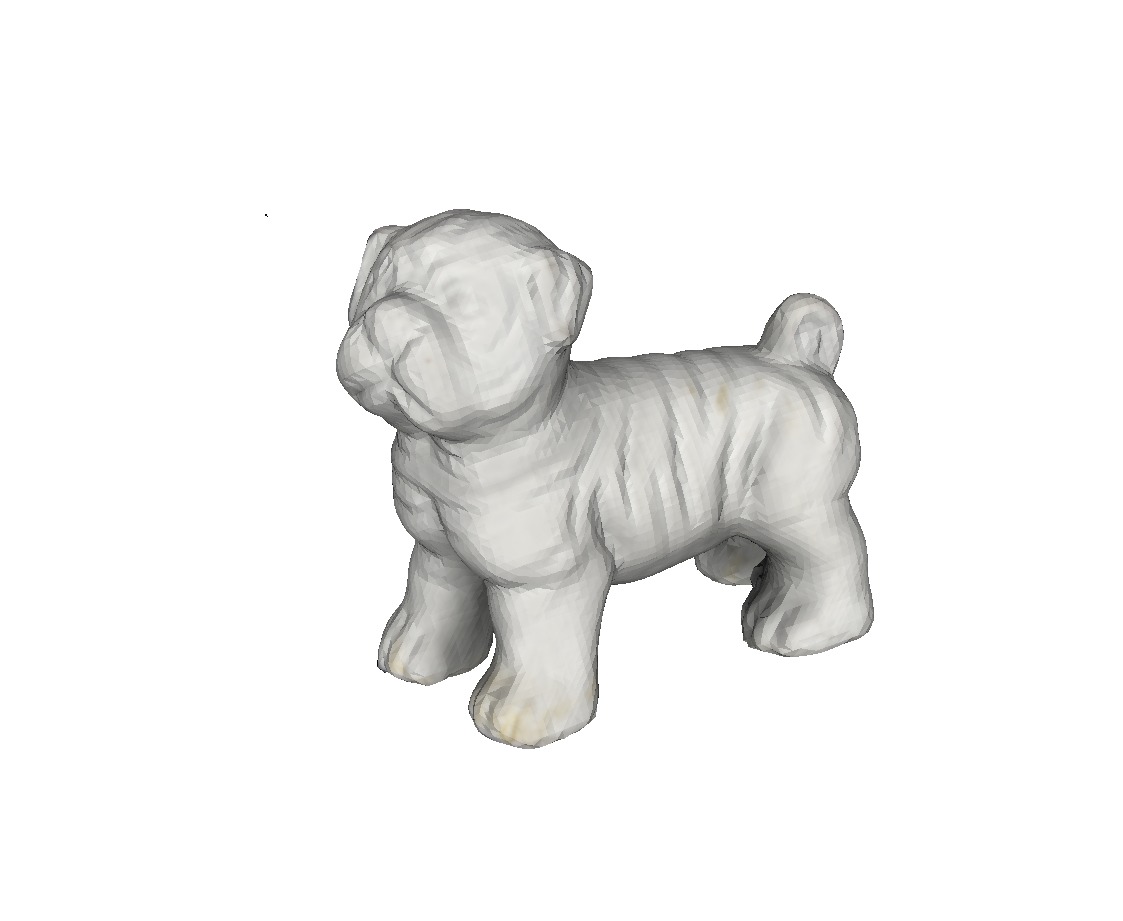} &
\includegraphics[trim=2cm 1.5cm 2cm 1.5cm,clip,width=.13\linewidth]{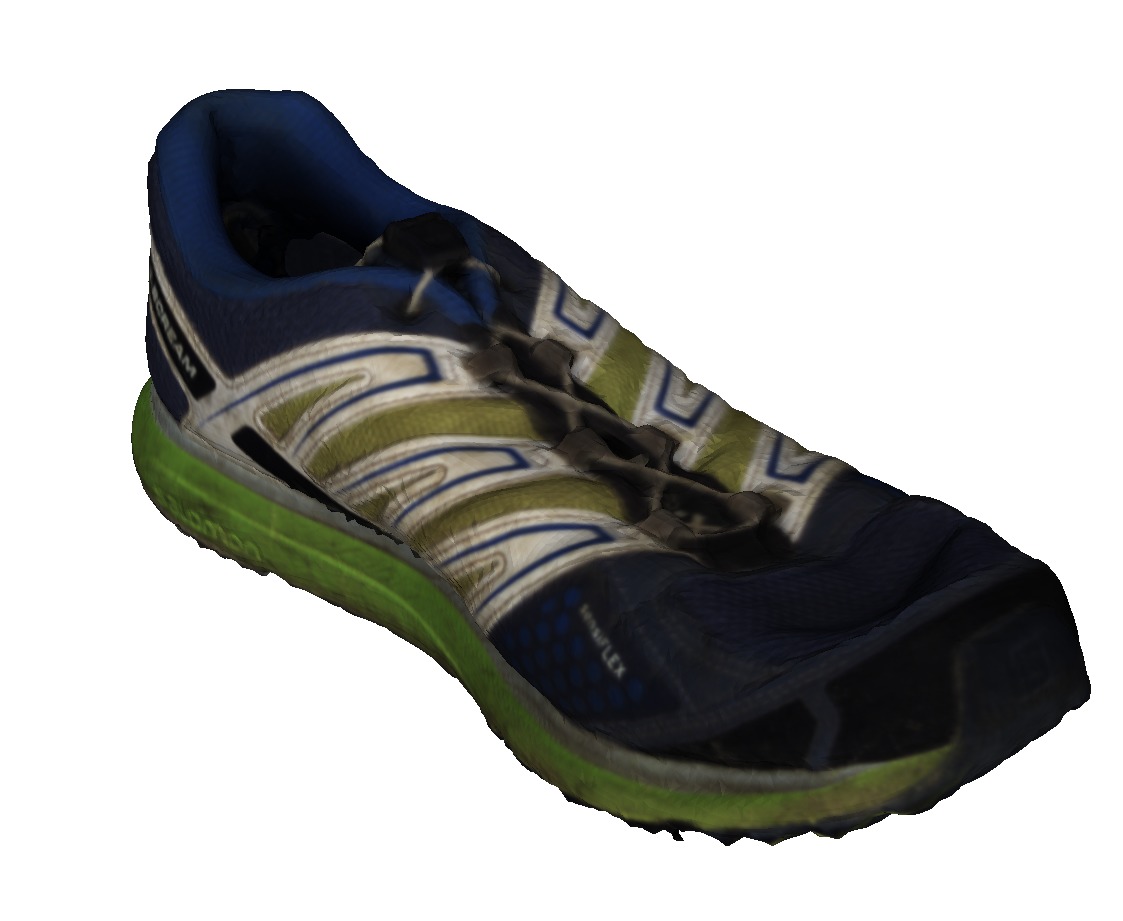} &
\includegraphics[trim=2cm 1.5cm 2cm 1.5cm,clip,width=.13\linewidth]{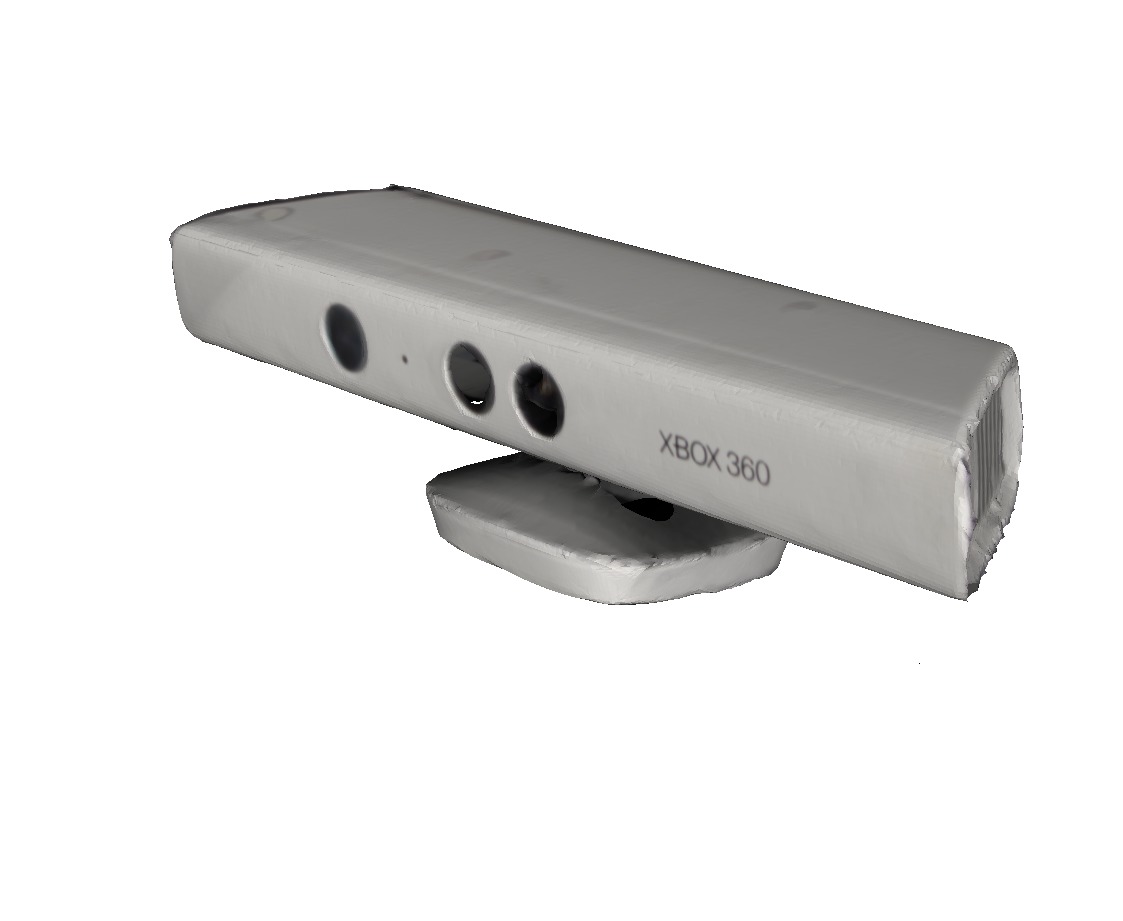} &
\includegraphics[trim=2cm 1.5cm 2cm 1.5cm,clip,width=.13\linewidth]{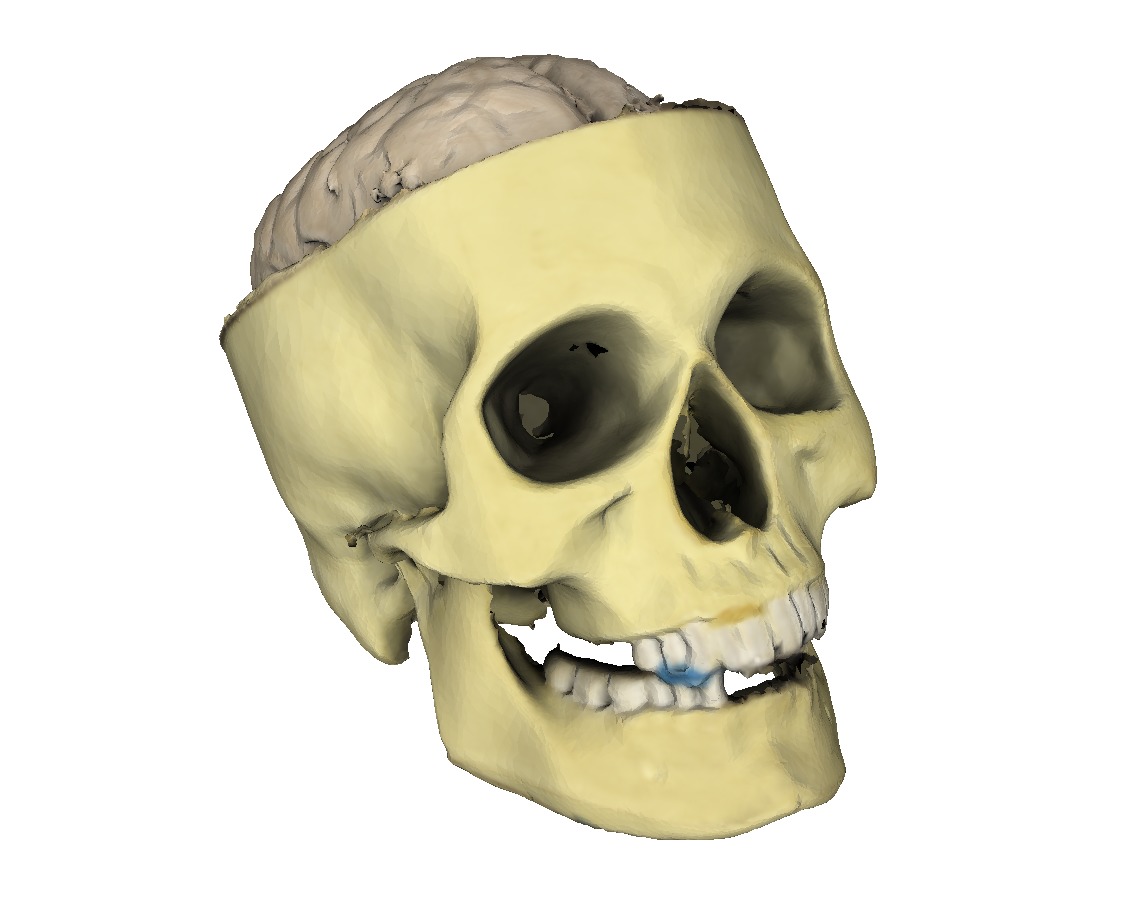} &
\includegraphics[trim=2cm 1.5cm 1cm 1.5cm,clip,width=.13\linewidth]{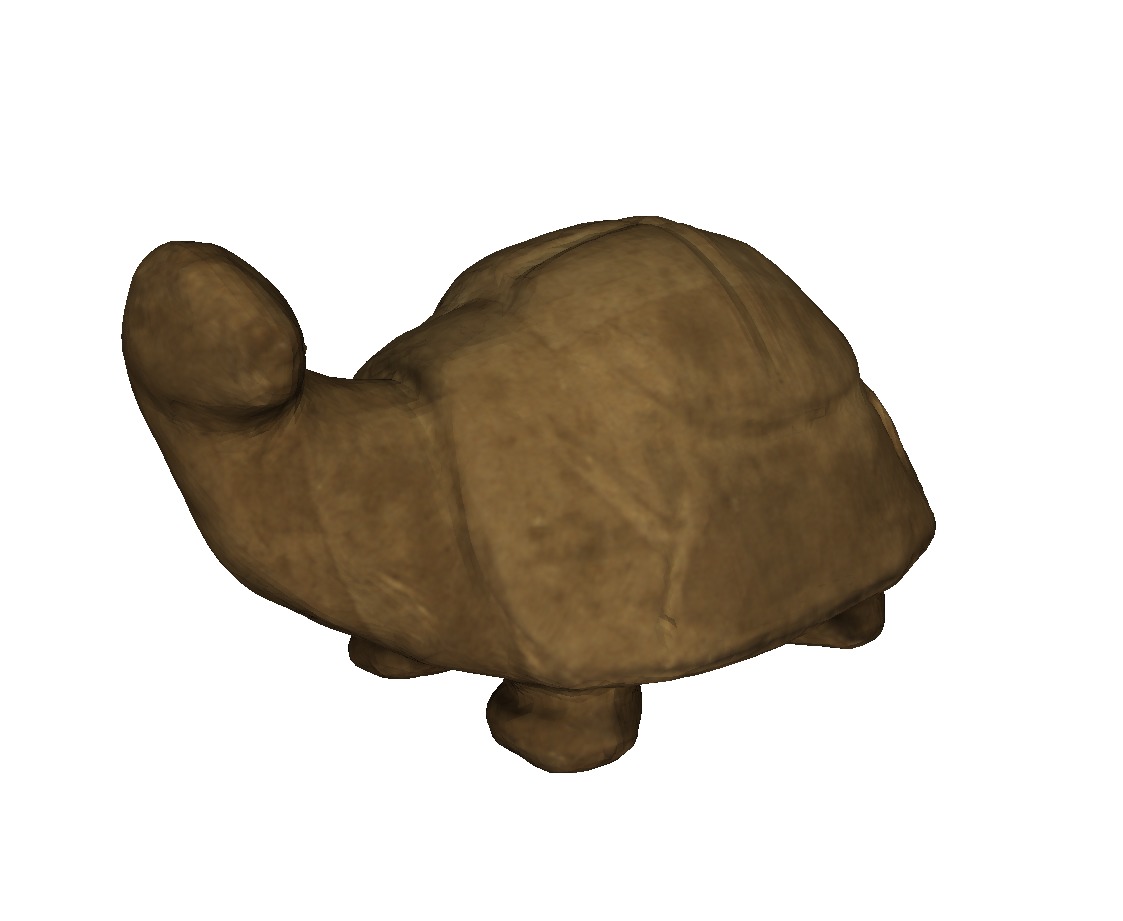} \\
\end{tabular}
\begin{tabular}{c@{\hskip 0.2in} c@{\hskip 0.2in} c@{\hskip 0.2in} c}
 lego (118) & watering can (287) & walkman (141) & cookie jar (187)\\
 \includegraphics[trim=2cm 1.5cm 2cm 1.5cm,clip,width=.13\linewidth]{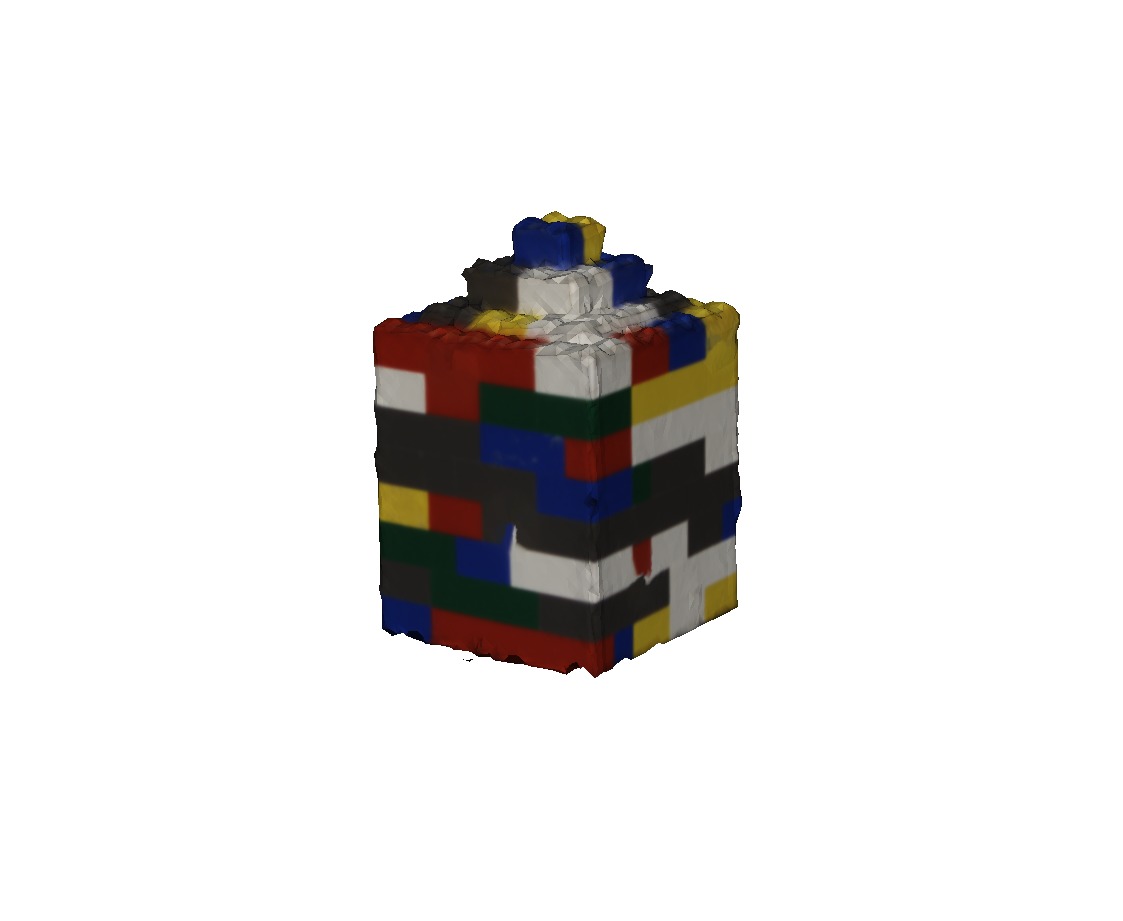} &
\includegraphics[trim=2cm 1.5cm 2cm 1.5cm,clip,width=.13\linewidth]{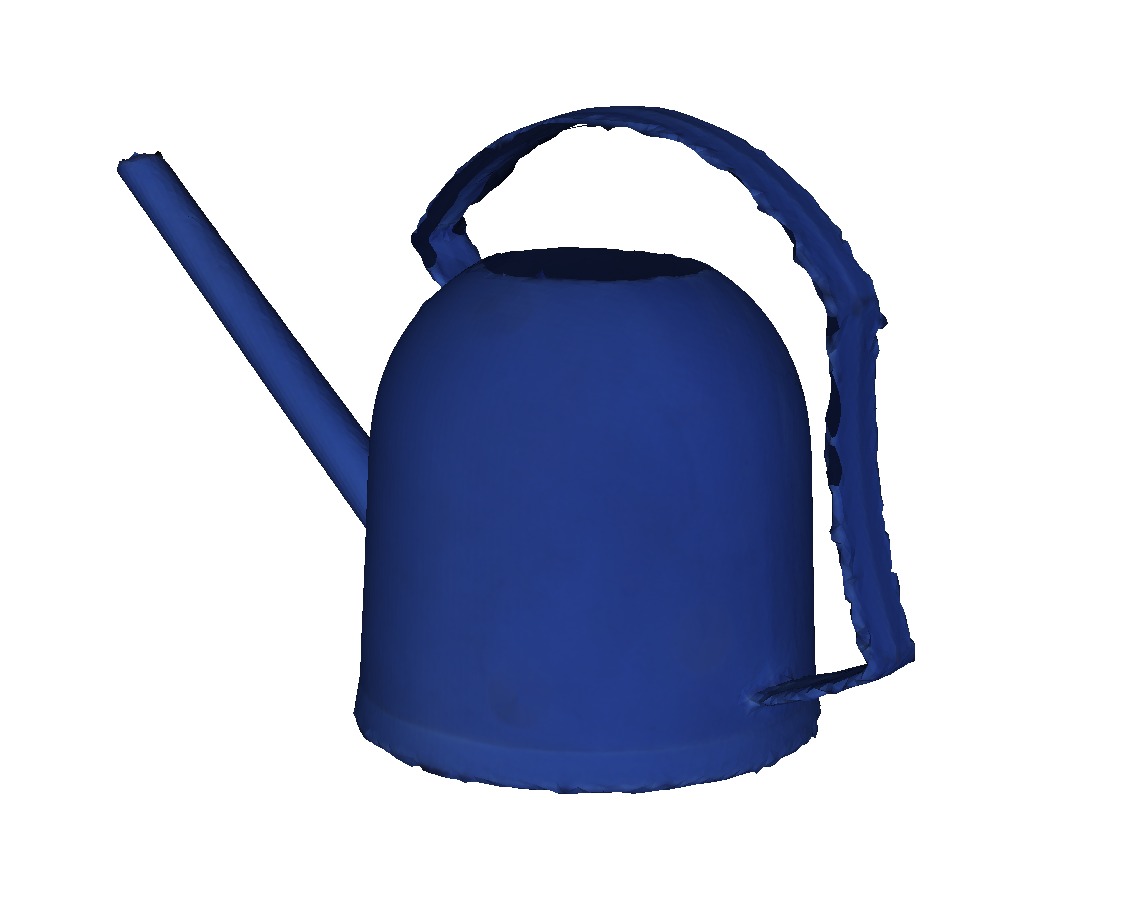} &
\includegraphics[trim=2cm 1.5cm 2cm 1.5cm,clip,width=.13\linewidth]{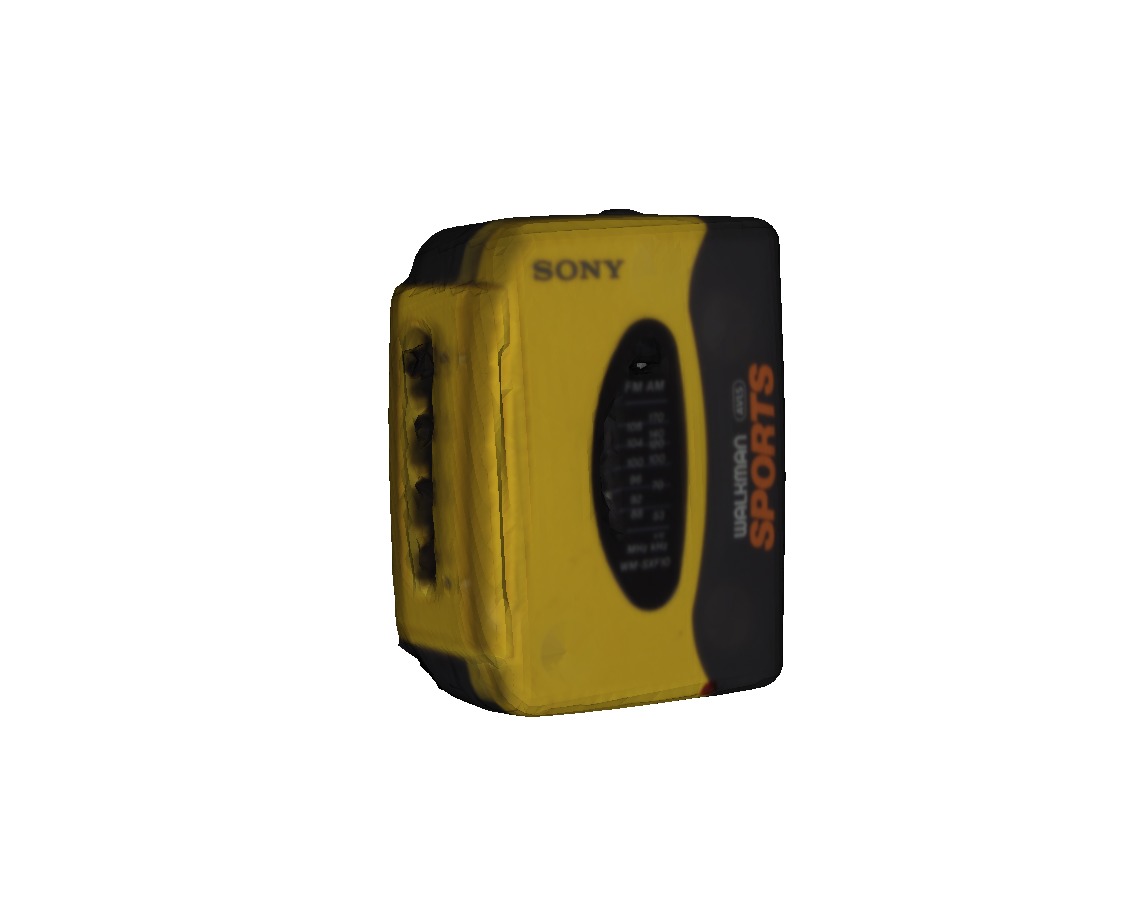} &
\includegraphics[trim=2cm 1.5cm 2cm 1.5cm,clip,width=.13\linewidth]{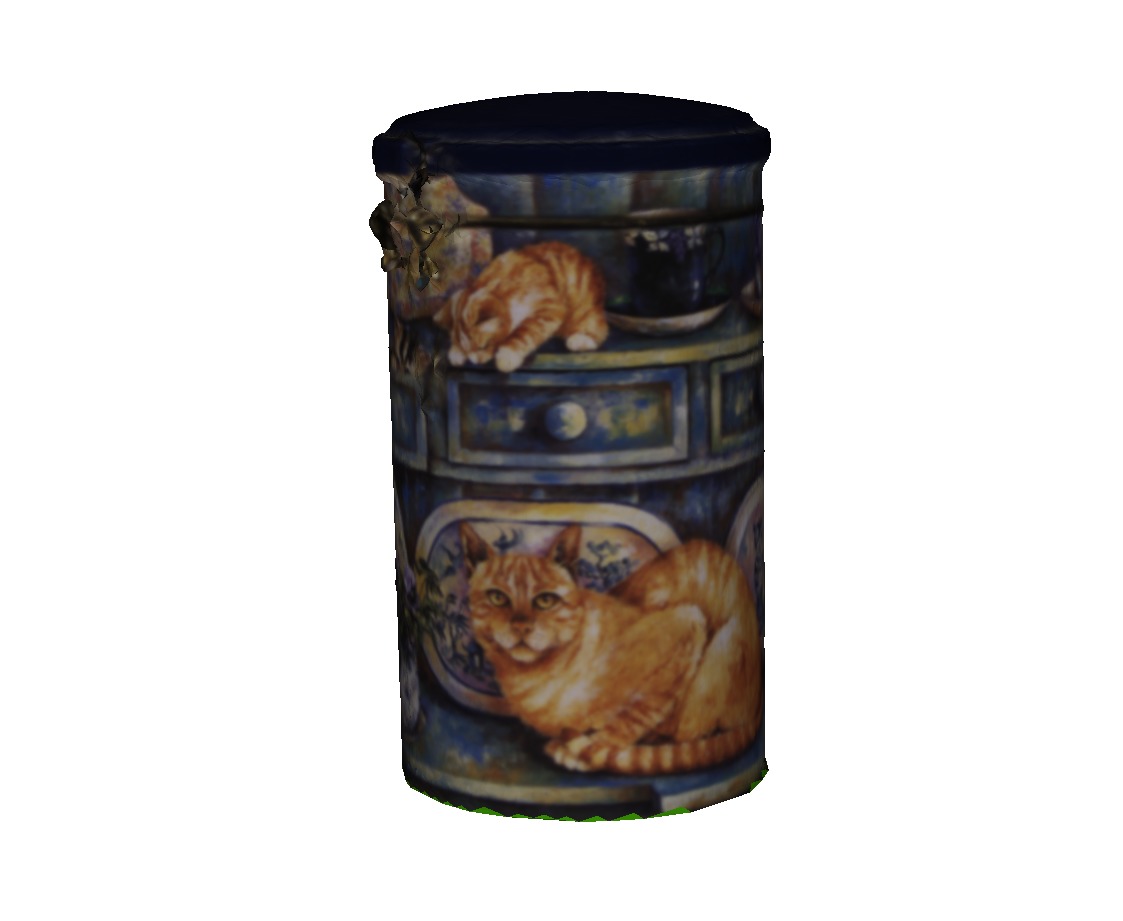}\\ 
\end{tabular}
\caption{Overview of the 11 objects in our dataset, with their maximum distance between two vertices in mm shown above.}
\label{fig:models}
\end{figure}

\subsection{Dataset statistics} 

We selected 11 different objects to obtain a wide variety of object geometries and appearance, as shown in fig.~\ref{fig:models}. To obtain a precise 3D model of each object in the database, each of them was scanned with a Creaform GoScan\texttrademark~handheld 3D scanner at a 1mm voxel resolution. The scans were manually cleaned using Creaform VxElements\texttrademark~to remove background and spurious vertices.

Overall, the dataset contains 297 sequences: 27 sequences for each object. The breakdown is the following: 12 sequences for \emph{stability} (4 viewpoints, 3 configurations: ``near'', ``far'', ``occluded''); 11 sequences for \emph{occlusion} (0\% to 75\% in 15\% increment for both horizontal and vertical occluders); and 4 sequences for \emph{interaction} (``rotation-only'', ``translation-only'', ``free-slow'', ``free-hard'').
It also contains high resolution textured 3D models for each object.

\section{Analyzing a deep 6-DOF tracker with our dataset}
\label{sec:improved_deep_6dof_tracker}

As a testbed to evaluate the relevance of the new dataset, we borrow the technique of Garon and Lalonde~\cite{garon-tvcg-17} who train a 6-DOF tracker using deep learning, but propose changes to their architecture and training methodology. We evaluate several variants of the network on our dataset and show that it can be used to accurately quantify the performance of a tracker in a wide variety of scenarios. 

\subsection{Training an object-specific tracker}
\label{sec:tracker-description}

\begin{figure}[!t]
\floatbox[{\capbeside\thisfloatsetup{capbesideposition={left},capbesidewidth=6cm}}]{figure}[\FBwidth]
{
\begin{tabular}{cc}
\centering
Input: $\mathbf{x}_\text{pred}$ & Input: $\mathbf{x}_\text{obs}$ \\
\noindent\rule{2cm}{.8pt} & \noindent\rule{2cm}{.8pt} \\
conv3-96 & conv3-96 \\
fire-48-96 & fire-48-96 \\
\end{tabular} 
\linebreak
\begin{tabular}{c}
concatenation \\
fire-96-384 \\
fire-192-768 \\
fire-384-768 \\
FC-500    \\
FC-6	 \\
\noindent\rule{2.5cm}{.8pt} \\
Output: $\mathbf{y}$ \\
\end{tabular}
}
{\caption[]{The deep learning architecture used to track 3D objects in this work, inspired by~\cite{garon-tvcg-17}. The notation ``conv$x$-$y$'' indicates a convolution layer of $y$ filters of dimension $x \times x$, ``fire-$x$-$y$'' indicates a ``fire'' module~\cite{SqueezeNet} which reduces the number of channels to $x$ and expands to $y$, and ``FC-$x$'' is a fully-connected layer of $x$ units. Each layer have a skip link similar to DenseNet~\cite{huang2017densely} and is followed by a max pooling $2 \times 2$ operation. We use a dropout of 50\% on the input connections to the FC-500 layer. All layers (except the last FC-6) have batch normalization and the ELU activation function~\cite{clevert-iclr-16}.}\label{fig:architecture}}
\end{figure}




We propose improvements over the previous work of~\cite{garon-tvcg-17} by adding 5 main changes: 2 to the network architecture, and 3 to the training procedure. The new proposed network architecture is shown in fig.~\ref{fig:architecture}. As in~\cite{garon-tvcg-17}, the network accepts two inputs: an image of the object rendered at its predicted position (from the previous timestamp in the video sequence) $\mathbf{x}_\text{pred}$, and an image of the observed object at the current timestamp $\mathbf{x}_\text{obs}$. The last layer outputs the 6-DOF (3 for translation, 3 for rotation in Euler angles) representing the pose change between the two inputs. We first replace convolution layers by the ``fire'' modules proposed in~\cite{SqueezeNet}. The second change, inspired by DenseNet~\cite{huang2017densely}, is to concatenate the input features of each layers to the outputs before being maxpooled. Our improvement requires the same runtime as~\cite{garon-tvcg-17}, which is 6 ms on a Nvidia GTX-970M. As in \cite{garon-tvcg-17}, the loss used is the MSE between the predicted and ground truth pose change. Note that we experimented with the reprojection loss~\cite{kendall2017geometric}, but found it did not help in our context.

We also propose changes to the training procedure of \cite{garon-tvcg-17}. Their approach consists in generating pairs of synthetic views of the object with random pose changes between them. To sample the random pose changes, they proposed to independently sample a random translation $t_{x,y,z} \sim U(20\text{mm},20\text{mm})$ and rotation $r_{\alpha,\beta,\gamma} \sim U(-10^\circ ,10^\circ)$ in Euler angle notation, with $U(a,b)$ referring to a uniform distribution on interval $[a,b]$. Doing so unfortunately biases the resulting pose changes. For example, small amplitude translations are quite unlikely to be generated (since this requires \emph{all} three translation components to be small simultaneously). Our first change is to sample a random translation \emph{vector} and \emph{magnitude} separately. The translation vector $\mathbf{v}_t$ is sampled in spherical coordinates $(\theta_t,\phi_t)$, where $\theta_t \sim U(-180^\circ,180^\circ)$ and $\phi_t = \cos^{-1}(x)$ with $x \sim U(-1,1)$. The translation magnitude $m_t$ is drawn from a Gaussian distribution $m_t \sim \mathcal{N}(0, \Delta t)$. The same process is repeated for rotations, where the rotation axis $\mathbf{v}_r$ and angle $m_r \sim \mathcal{N}(0, \Delta r)$ are sampled similarly. Here, we intentionally parameterize the translation magnitude $m_t$ and rotation angle $m_r$ distributions with $\Delta t$ and $\Delta r$, since the range of these parameters may influence the behavior of the network. Our second change is to downsample the depth channel to better match the resolution of the Kinect V2. Our third change consists in a data augmentation method for RGBD images where we randomly set a modality (depth or RGB) to zero during training, which has the effect of untangling the features of both modalities. With these changes, we can now rely purely on synthetic data to train the network (in~\cite{garon-tvcg-17} a set of real frames was required to fine-tune the network). 

\subsection{Training a generic tracker}
\label{sec:generic-tracker}

To train a generic 6-DOF object tracker, we experimented with two ways of generating a training dataset, using the same network architecture, loss, and training procedure described in sec.~\ref{sec:tracker-description}. First, we generate a training set of images that contain all 11 objects from our dataset, as well as 30 other objects. These other objects, downloaded from 3D Warehouse\footnote{Available at: \url{https://3dwarehouse.sketchup.com}.} and from ``Linemod''~\cite{hinterstoisser2012model}, show a high diversity in geometry and texture and are roughly of the same size. We name the network trained on this dataset the ``multi-object'' network. Second, we generate a training set of images that contain only the 30 \emph{other} objects---the actual objects to track are \emph{not} included. We call this network ``generic'', since it never saw any of the objects in our dataset during training. Note that all these approaches require the 3D model of the object to track at test time, however.

\begin{table}[!t]
	\centering
	\tiny

	\sisetup{detect-weight=true,detect-inline-weight=math}
	\begin{tabular}{rr *{3}{S[table-format=\sizeA]} c@{\hskip 0.1in} cc *{3}{S[table-format=\sizeA]}}
	\multicolumn{1}{l}{\textbf{Stability Scenario}} && \multicolumn{3}{c}{\textbf{Translation (mm/frame)}} & & & & \multicolumn{3}{c}{\textbf{Rotation (degree/frame)}} \\
	\cmidrule(lr){3-5}
	\cmidrule(ll){9-11}
	&& \mc{Near} & \mc{Far} & \mc{Occluded} & & & & \mc{Near} & \mc{Far} & \mc{Occluded}	\\
	\cmidrule(lr){3-5}
	\cmidrule(ll){9-11}
 \multirow{5}{*}{\rotatebox{90}{$\Delta t$}} & 10 & \boldentry{\sizeA}{0.42} & \boldentry{\sizeA}{0.53} & \boldentry{\sizeA}{0.48} && \parbox[t]{2mm}{\multirow{5}{*}{\rotatebox[origin=c]{90}{$\Delta r$}}} & 15 & \boldentry{\sizeA}{0.35} & \boldentry{\sizeA}{0.51} & \boldentry{\sizeA}{0.63} \\
 & 20 & 0.62 & 0.77 & 0.71 &&& 20 & 0.55 & 0.68 & 0.88 \\ 
 & 30 & 0.72 & 0.84 & 0.89 &&& 25 & 0.53 & 0.66 & 0.97 \\ 
 & 40 & 0.75 & 0.89 & 1.00 &&& 30 & 0.61 & 0.74 & 1.04 \\ 
 & 50 & 0.95 & 1.03 & 1.16 &&& 35 & 0.68 & 0.87 & 1.10 \\ 
	\end{tabular}
	\\*[1em]


	\begin{tabular}{rr *{6}{S[table-format=\sizeC]} c@{\hskip 0.1in} cc  *{6}{S[table-format=\sizeC]}}
	\multicolumn{1}{l}{\textbf{Occlusion scenario }}&& \multicolumn{6}{c}{\textbf{Translation (mm)}} & & & & \multicolumn{6}{c}{\textbf{Rotation (degrees)}} \\
	\cmidrule(lr){3-8}
	\cmidrule(ll){12-17}
	Occlusion \% & & \mc{0} & \mc{15} & \mc{30} & \mc{45} & \mc{60} & \mc{75} & & & & \mc{0} & \mc{15} & \mc{30} & \mc{45} & \mc{60} & \mc{75} \\
	\cmidrule(lr){3-8}
	\cmidrule(ll){12-17}
 \parbox[t]{2mm}{\multirow{5}{*}{\rotatebox[origin=c]{90}{$\Delta t$}}} & 10 & 14.8 & 12.5 & 13.1 & \boldentry{\sizeC}{15.5} &  \boldentry{\sizeC}{20.0} &  \boldentry{\sizeC}{25.3} && \parbox[t]{2mm}{\multirow{5}{*}{\rotatebox[origin=c]{90}{$\Delta r$}}} & 15 & \boldentry{\sizeC}{2.1} & \boldentry{\sizeC}{2.9} & \boldentry{\sizeC}{5.4} & \boldentry{\sizeC}{8.1} & \boldentry{\sizeC}{15.7} &  \boldentry{\sizeC}{26.0} \\ 
& 20 & \boldentry{\sizeC}{7.5}  & \boldentry{\sizeC}{7.6}  & \boldentry{\sizeC}{12.5} & \boldentry{\sizeC}{15.5} & 22.3 & 29.5 &&& 20 & 3.3 & 4.8 & 7.5 & 9.9 & 17.2 & 48.0 \\
& 30 & 11.0 & 11.5 & 17.4 & 21.5 & 26.6 & 33.9 &&& 25 & 3.3 & 4.8 & 8.7 & 16.8 & 30.6 & 41.1 \\ 
& 40 & 12.7 & 14.8 & 17.9 & 26.0 & 36.0 & 68.1 &&& 30 & 2.7 & 3.9 & 6.9 & 12.3 & 26.9 & 62.9 \\ 
& 50 & 10.4 & 11.1 & 17.7 & 30.6 & 43.8 & 73.7 &&& 35 & 3.2 & 4.6 & 9.1 & 16.1 & 36.7 & 66.1 \\ 
	\end{tabular}
	\\*[1em]
	\begin{tabular}{rr *{4}{S[table-format=4.3]} c@{\hskip 0.05in} cc *{4}{S[table-format=4.3]}}
	\multicolumn{1}{l}{\textbf{Interaction scenario}}&& \multicolumn{4}{c}{\textbf{Translation (mm)}} & & & & \multicolumn{4}{c}{\textbf{Rotation (degrees)}} \\
	\cmidrule(lr){3-6}
	\cmidrule(ll){10-13}
	Speed (per frame)&& \mc{(0, 10]} & \mc{(10, 20]} & \mc{(20, 30]} & \mc{(30, 40]} & & & &
	        \mc{(0, 4]} & \mc{(4, 8]} & \mc{(8, 12]} & \mc{(12, 16]}\\
	\cmidrule(lr){3-6}
	\cmidrule(ll){10-13}
\parbox[t]{2mm}{\multirow{5}{*}{\rotatebox[origin=c]{90}{$\Delta t$}}}& 10 & 10.1 & 11.8 & 16.4 & 18.8 && \parbox[t]{2mm}{\multirow{5}{*}{\rotatebox[origin=c]{90}{$\Delta r$}}} & 15 & \boldentry{4.3}{2.9} & \boldentry{4.3}{3.8} & \boldentry{4.3}{4.2} & \boldentry{4.3}{4.4}\\ 
& 20 & \boldentry{4.3}{6.5}  & \boldentry{4.3}{8.7}  & 11.0 & 18.0 &&& 20 & 3.7 & 4.7 & 4.9 & 5.0\\ 
& 30 & 9.7  & 9.9  & 11.5 & 9.5  &&& 25 & 5.7 & 5.9 & 5.8 & 6.1\\ 
& 40 & 10.8 & 11.4 & 11.7 & \boldentry{4.3}{6.6}  &&& 30 & 4.3 & 4.6 & 4.7 & 4.7\\ 
& 50 & 10.5 & 11.4 & \boldentry{4.3}{10.9} & 7.2  &&& 35 & 4.7 & 4.8 & 5.4 & 5.1\\ 
\\
	&& \multicolumn{4}{c}{(a) Impact of $\Delta t$ on $\delta_\mathbf{t}$} & & & & \multicolumn{4}{c}{(b) Impact of $\Delta r$ on $\delta_\mathbf{R}$} \\
	\end{tabular}
	\caption{Applying our evaluation methodology for determining the best range of translations $\Delta t$ and rotations $\Delta r$ for generating synthetic data when training a deep 6-DOF tracker. We show (a) the impact of varying $\Delta t$ on the error $\delta_\mathbf{t}$, and (b) the impact of varying $\Delta r$ on the error $\delta_\mathbf{R}$ for all three scenarios (from top to bottom: \emph{stability}, \emph{occlusion}, and \emph{interaction}). } 
	\label{tab:methodology-single-object}
\end{table}

\section{Experiments}
\label{sec:evaluation}

In this section, we perform an exhaustive evaluation of the various approaches presented in sec.~\ref{sec:improved_deep_6dof_tracker} using our novel dataset and framework proposed in sec.~\ref{sec:evaluation_methodology}. First, we analyze the impact of varying the training data generation hyper-parameters $\Delta t$ and $\Delta r$ for the object-specific case. Then, we proceed to compare our object-specific, ``multi-object'', and ``generic'' trackers with two existing methods: Garon and Lalonde~\cite{garon-tvcg-17} and Tan et al.~\cite{tan-iccv-15}. 

\subsection{Analysis to training data generation parameters}
\label{subsec:analysis}

We now apply the evaluation methodology proposed in sec.~\ref{sec:evaluation_methodology} on the method presented above and evaluate the influence of the $\Delta r$ and $\Delta t$ hyper-parameters on the various metrics and sequences from our dataset. We experiment by varying $\Delta t \in \{10, 20, 30, 40, 50\}\text{mm}$ and $\Delta r \in \{15, 20, 25, 30, 35\}^\circ$ one at a time (the other parameter is kept at its lowest value). For each of these parameters, we synthesize 200,000 training image pairs per object using \cite{garon-tvcg-17} and the modifications proposed in sec.~\ref{sec:tracker-description}. We then train a network for each object, for each set of parameters, and evaluate each network on our dataset. A subset of the results of this analysis is shown in tab.~\ref{tab:methodology-single-object}. Note that, for the \emph{interaction} scenario, the ``free-hard'' sequences (sec.~\ref{subsec:free-form_interaction}) were left out since they are much harder than the others and would bias the results. In particular, we show the impact that varying $\Delta t$ has on $\delta_\mathbf{t}$, as well as that of varying $\Delta r$ has on $\delta_\mathbf{R}$ for all 3 scenarios. Here, we drop the parentheses for the $\delta_{\{\mathbf{t}, \mathbf{R}\}}$ error metrics for ease of notation (see sec.~\ref{sec:evaluation_methodology} for the definitions). The figure reveals a clear trend: increasing $\Delta r$ (tab.~\ref{tab:methodology-single-object}-(b)) systematically results in worse performance in rotation. This is especially visible for the high occlusion cases (45\% and 60\%), where the rotation error $\delta_\mathbf{R}$ increases significantly as a function of $\Delta r$. The situation is not so simple when $\Delta t$ is increased (tab.~\ref{tab:methodology-single-object}-(a)). Indeed, while increasing $\Delta t$ negatively impacts $\delta_\mathbf{t}$ in the \emph{stability} and \emph{occlusion} scenarios, performance actually \emph{improves} when the object speed is higher, as seen in the \emph{interaction} scenario. Therefore, to achieve a good balance between stability and accuracy at higher speeds, a value of $\Delta t = 30\text{mm}$ seems to be a good trade-off. The remainder of the plots for this analysis, as well as plots evaluating the impact of the resolution of the crop and the size of the bounding box w.r.t the object are shown in the supplementary material.

\begin{table}[t!]
	\centering
	\tiny

	\sisetup{detect-weight=true,detect-inline-weight=math}
	\begin{tabular}{l *{3}{S[table-format=\sizeA]} *{3}{S[table-format=\sizeA]}}
	\textbf{Stability scenario}	& \multicolumn{3}{c}{\textbf{Translation (mm/frame)}} & \multicolumn{3}{c}{\textbf{Rotation (degree/frame)}} \\
	\cmidrule(lr){2-4}
	\cmidrule(ll){5-7}
	& \mc{Near} & \mc{Far} & \mc{Occluded} & \mc{Near} & \mc{Far} & \mc{Occluded}	\\
	\cmidrule(lr){2-4}
	\cmidrule(ll){5-7}
	Ours specific & 0.56 & 0.68 & 0.72 & \boldentry{\sizeA}{0.52} & \boldentry{\sizeA}{0.59} & \boldentry{\sizeA}{0.76} \\ 
	Ours multi-object & \boldentry{\sizeA}{0.38} & \boldentry{\sizeA}{0.41} & \boldentry{\sizeA}{0.57} & 0.69 & 0.79 & 1.09 \\ 
	Ours generic & 0.72 & 0.75 & 1.19 & 0.95 & 0.98 & 1.67 \\ 
	Garon and Lalonde [1] & 0.93 & 1.06 & 1.24 & 1.13 & 1.23 & 1.49 \\ 
	Tan et al [5] & 1.20 & 1.31 & 1.53 & 1.30 & 1.44 & 1.92 \\ 
	\end{tabular}
	\\*[1em]


	\begin{tabular}{l *{6}{S[table-format=\sizeB]}  *{6}{S[table-format=\sizeB]}}
	\textbf{Occlusion scenario} & \multicolumn{6}{c}{\textbf{Translation (mm)}} & \multicolumn{6}{c}{\textbf{Rotation (degrees)}} \\
	\cmidrule(lr){2-7}
	\cmidrule(ll){8-13}
	Occlusion \% & \mc{0} & \mc{15} & \mc{30} & \mc{45} & \mc{60} & \mc{75} & \mc{0} & \mc{15} & \mc{30} & \mc{45} & \mc{60} & \mc{75}	\\
	\cmidrule(lr){2-7}
	\cmidrule(ll){8-13}
Ours specific & 7.4 & 9.8 & \boldentry{\sizeB}{11.5} & \boldentry{\sizeB}{12.5} & 15.5 & 24.2 & \boldentry{\sizeB}{3.6} & \boldentry{\sizeB}{5.9} & \boldentry{\sizeB}{7.9} & \boldentry{\sizeB}{10.0} & \boldentry{\sizeB}{12.6} & 22.2 \\ 
Ours multi-object & 23.3 & 16.9 & 14.6 & 14.3 & \boldentry{\sizeB}{13.2} & \boldentry{\sizeB}{13.4} & 4.0 & 8.6 & 12.3 & 12.1 & 14.7 & \boldentry{\sizeB}{15.4} \\ 
Ours generic & \boldentry{\sizeB}{6.7} & 11.1 & 18.9 & 25.9 & 34.4 & 47.4 & 5.3 & 8.4 & 16.1 & 26.8 & 43.7 & 50.6 \\ 
Garon and Lalonde [1] & 7.4 & 11.2 & 18.9 & 26.8 & 38.1 & 55.0 & 5.3 & 8.8 & 17.7 & 28.2 & 41.7 & 49.8 \\ 
Tan et al [5] & 8.2 & \boldentry{\sizeB}{8.5} & 15.9 & \mc{138} & \mc{186} & \mc{213} & 4.0 & 7.4 & 33.1 & 70.3 & 89.5 & 88.0 \\ 
	\end{tabular}
	\\*[1em]
	\begin{tabular}{l *{8}{S[table-format=\sizeC]}{S[table-format=3.1]}}
	\textbf{Interaction scenario} & \multicolumn{4}{c}{\textbf{Translation (mm)}} & \multicolumn{4}{c}{\textbf{Rotation (degrees)}} \\
	\cmidrule(lr){2-5}
	\cmidrule(ll){6-9}
	Speed (per frame) & \mc{(0, 12.5]} & \mc{(12.5, 25]} & \mc{(25, 37.5]} & \mc{(37.5, 50]} &
	        \mc{(0, 19]} & \mc{(19, 37]} & \mc{(37, 56]} & \mc{(56, 75]} & \mc{Fail} \\
	\cmidrule(lr){2-5}
	\cmidrule(ll){6-9}
Ours specific & 8.2 & 10.3 & 11.1 & 13.4 & \boldentry{\sizeC}{3.7} & \boldentry{\sizeC}{5.8} & \boldentry{\sizeC}{3.6} & \boldentry{\sizeC}{5.8} & \boldentry{3.1}{37} \\ 
Ours multi-object & 22.1 & 27.3 & 26.0 & 41.9 & 6.0 & 8.6 & 2.9 & 6.1 & 127 \\ 
Ours generic & 9.3 & 9.9 & 11.7 & 13.4 & 6.3 & 6.8 & 8.8 & 7.0 & 38 \\ 
Garon and Lalonde [1] & 9.5 & 10.2 & \boldentry{\sizeC}{10.3} & \boldentry{\sizeC}{12.4} & 7.8 & 9.5 & 13.4 & 11.8 & 53 \\ 
Tan et al [5] & \boldentry{\sizeC}{8.1} & \boldentry{\sizeC}{8.5} & 10.7 & 67.1 & 4.5 & 6.0 & 8.1 & 10.1 & 86 \\ 
	\end{tabular}
	\caption{Comparison of our networks with the previous work of \cite{garon-tvcg-17} and \cite{tan-iccv-15}. Our ``object-specific'' networks outperform the state of the art in almost all scenarios, and performs remarkably well at predicting the rotation. Our ``generic'' tracker shows great promise: although not as good as the ``object-specific'' version, it results in slightly lower error compared with \cite{garon-tvcg-17}, even if it has not seen any of these objects during training. See the supplementary video for a visual qualitative comparison of the trackers.}
	\label{tab:results}
\end{table}

\subsection{Comparison with previous work}

Our trackers yields a $1.7\text{mm}/0.6^\circ$ error on the 4 sequences of \cite{choi2013rgb} which is slightly above \cite{tan-iccv-15} who obtain $0.81\text{mm}/0.37^\circ$. However, as reported in tab.~\ref{tab:results}, more interesting differences between the trackers can be observed when using our new dataset. We compare with object-specific versions of the work of Garon and Lalonde~\cite{garon-tvcg-17} as well as the Random Forest approach of Tan et al.~\cite{tan-iccv-15}. For \cite{garon-tvcg-17}, we use the training parameters reported in their paper. For our trackers, the $\Delta m$ and $\Delta r$ hyper-parameters were obtained with leave-one-out cross-validation to ensure no training/test overlap. As before, the ``free-hard'' sequences were left out for the \emph{interaction} experiments. 

Overall, as can be observed in tab.~\ref{tab:results}, the proposed deep learning methods perform either on par or better than the previous work. The ``object-specific'' networks outperform almost all the other techniques, except for the case of translational error in the interaction scenario. It performs remarkably well at predicting rotations, and is on par with the other methods for translation. In comparison, \cite{tan-iccv-15} performs well at low occlusions, but fails when the occlusion level is 30\% or greater (particularly in rotation). \cite{garon-tvcg-17} shows improved robustness to occlusions, but still achieves high rotation errors at 45\% occlusion, and is also much less stable (esp. in rotation) than our ``object-specific'' networks. Interestingly, our ``generic'' tracker, which has seen none of these objects in training, performs similarly to the previous works that were trained specifically on these objects. Indeed, it shows a stability, robustness to occlusions and behavior at higher speeds that is similar to \cite{garon-tvcg-17} and \cite{tan-iccv-15}, demonstrating that learning generic features that are useful for tracking objects can be achieved. Finally, we use the ``free-hard'' \emph{interaction} sequences to count the number of times the tracking is lost (sec.~\ref{subsec:stability}). In this case, the ``object-specific'' and ``generic'' networks outperforms the other methods. Qualitative videos showing side-by-side comparisons of these methods are available in the supplementary material.

\section{Discussion}
\label{sec:discussion}

The recent evolution in 6-DOF tracking performance on the popular dataset of Choi et al.~\cite{choi2013rgb} highlights the need for a new dataset containing real data and more challenging scenarios. In this paper, we provide such a dataset, which we hope will spur further research in the field. Our dataset contains 297 sequences containing 11 objects of various shapes and textures. The sequences are grouped into 3 scenarios: \emph{stability}, \emph{occlusion}, and \emph{interaction}. The dataset and companion evaluation code is released publicly\footnote{\url{http://www.jflalonde.ca/projects/6dofObjectTracking}}. Additionally, we build on the framework of \cite{garon-tvcg-17} with an improved architecture and training procedure which allows the network to learn purely from synthetic data, yet generalize well on real data. In addition, the architecture allows for training on multiple objects and test on \emph{different} objects it has never seen in training. To the best of our knowledge, we are the first to propose such a generic learner for the 6-DOF object tracking task. Finally, our approach is extensively compared with recent work and is shown to achieve better performance. 

A current limitation is that the Vicon markers must be removed in a post-processing step, which may leave some artifacts behind. While the markers are very small (3mm) and the resulting marker-free images have low error (see fig.~\ref{fig:markers}), there might still be room for improvement. Finally, our ``generic'' tracker is promising, but it still does not perform quite as well as ``object-specific'' models, especially for rotations. In addition, a 3D model of the object is still required at test time, so exploring how this constraint can be removed would make for an exciting future research direction.

\section*{Acknowledgements}
The authors wish to thank Jonathan Gilbert for his help with data acquisition and Sylvain Comtois for the Vicon setup. This work was supported by the NSERC/Creaform Industrial Research Chair on 3D Scanning: CREATION 3D. We gratefully acknowledge the support of Nvidia with the donation of the Tesla K40 and Titan X GPUs used for this research.

\bibliographystyle{splncs}
\bibliography{egbib}


\end{document}